\documentclass{article}

\PassOptionsToPackage{numbers, compress}{natbib}

\usepackage[preprint]{neurips_2025}
 \usepackage{amsmath}
\usepackage{fontawesome5}
\usepackage{enumitem}
\usepackage{wrapfig}
\usepackage{svg}
\usepackage{forest}
\usepackage{makecell}
\usepackage[table,xcdraw]{xcolor}
\usepackage{multirow}
\usepackage[utf8]{inputenc} 
\usepackage[T1]{fontenc}    
\usepackage{hyperref}       
\usepackage{url}            
\usepackage{booktabs}       
\usepackage{amsfonts}       
\usepackage{nicefrac}       
\usepackage{microtype}      
\usepackage{xcolor}         
\usepackage[most]{tcolorbox}
\usepackage{fontawesome5}
\usepackage{hyperref}
\usepackage{array} 
\usepackage{listings}
\usepackage{graphicx}  
\usepackage{subcaption} 
\usepackage{float}      
\usepackage{fontawesome5}
\usepackage{booktabs}
\usepackage{graphicx}
\usepackage{colortbl}
\usepackage[table]{xcolor}

\usepackage{placeins}

\usepackage{enumitem}  
\usepackage{tcolorbox} 
\usepackage{amssymb}   
\tcbuselibrary{breakable}
\usepackage{pifont}
\definecolor{lightblue}{RGB}{235,245,255} 
\definecolor{liautoblue}{RGB}{71,111,182} 
\definecolor{textred}{RGB}{128,0,0}
\usepackage{titlesec}
\titleformat{\section}
  {\large\bfseries\color{liautoblue}}{\thesection}{1em}{}
\titleformat{\subsection}
  {\bfseries\color{liautoblue}}{\thesubsection}{1em}{}
  \titleformat{\subsubsection}
  {\bfseries\color{liautoblue}}{\thesubsubsection}{1em}{}
\usepackage[most]{tcolorbox}
\newtcolorbox{liautoabstract}{
    colback=lightblue,
    colframe=white,
    boxrule=0pt,
    arc=2mm,
    left=4mm,
    right=4mm,
    top=5mm,
    bottom=5mm,
    enhanced, 
    before upper={\setlength{\parindent}{0pt}} 
}
\usepackage[most]{tcolorbox} 

\newtcolorbox{stepTitle}[1]{
    enhanced,
    colback=gray!5,    
    colframe=black!50,  
    boxrule=-1pt,
    arc=0mm,            
    left=2mm, right=2mm, top=1mm, bottom=1mm,
    fontupper=\small,  
    title=#1
}

\tcbset{
    stepbox/.style={
        enhanced,
        colback=gray!20,         
        colframe=gray!50,        
        boxrule=0.5pt,
        title={#1},              
        fonttitle=\bfseries,     
        left=2mm, right=2mm, top=1mm, bottom=1mm
    }
}

\newtcolorbox[auto counter]{case}[2][]{ 
    enhanced,
    colback=gray!5,
    colframe=black!70,
    coltitle=white,
    fonttitle=\bfseries\sffamily,
    fontupper=\small,
    arc=1.5mm,
    boxrule=0.5pt,
    title=Case study \thetcbcounter: #2, 
    left=1mm, right=1mm, top=2mm, bottom=2mm,
    label type=case, 
    #1               
}

\newtcolorbox{toolbox}[1]{
    enhanced,                 
    colback=gray!5,           
    colframe=black!70,        
    coltitle=white,           
    fonttitle=\bfseries\sffamily,
    fontupper=\small,
    arc=1.5mm,                
    boxrule=0.5pt,            
    title=#1,                 
    left=1mm, right=1mm, top=2mm, bottom=2mm
}

\usepackage{fancyhdr}
\usepackage{graphicx} 
\hypersetup{
    colorlinks=true,
    urlcolor=liautoblue,
}

\title{Mind DeepResearch Technical Report}

\author{%
  MindDR Team, Li Auto Inc   \\
}

\begin{document}

\maketitle

\begin{liautoabstract} 
We present \textbf{Mind DeepResearch (MindDR)}, an efficient multi-agent deep research framework that achieves leading performance with only \textasciitilde30B-parameter models through a meticulously designed data synthesis and multi-stage training pipeline. The core innovation of MindDR lies in a collaborative three-agent architecture (Planning Agent, DeepSearch Agent, and Report Agent) and a four-stage agent-specialized training pipeline comprising SFT cold-start, Search-RL, Report-RL and preference alignment. With this regime, MindDR  demonstrates competitive performance even with \textasciitilde30B-scale models. Specifically, MindDR achieves 45.7\% on BrowseComp-ZH, 42.8\% on BrowseComp, 46.5\% on WideSearch, 75.0\% on xbench-DS, and 52.5 on DeepResearch Bench, outperforming comparable-scale open-source agent systems and rivaling larger-scale models. MindDR has been deployed as an online product in Li Auto. Furthermore, we introduce \textbf{MindDR Bench}, a curated benchmark of 500 real-world Chinese queries from our internal product user interactions, evaluated through a comprehensive multi-dimensional rubric system rather than relying on a single RACE metric. On MindDR Bench, MindDR achieves a state-of-the-art score of 51.8.

\vspace{3mm}
    {\color{liautoblue!30}\rule{\linewidth}{0.5pt}} 
    \vspace{2mm}

    \small 
    \renewcommand{\arraystretch}{1.3} 
\begin{tabular}{@{} l l @{}}
        {\faCalendar*} & \textbf{Last Update Date:} April 14, 2026 \\
        {\color{liautoblue}\faEnvelope} & \textbf{Correspondence:} {yangsheng3@lixiang.com} \\
    \end{tabular}\end{liautoabstract}

\section{Introduction}


\begin{figure}[t]
    \centering
    \includegraphics[width=0.97\linewidth]{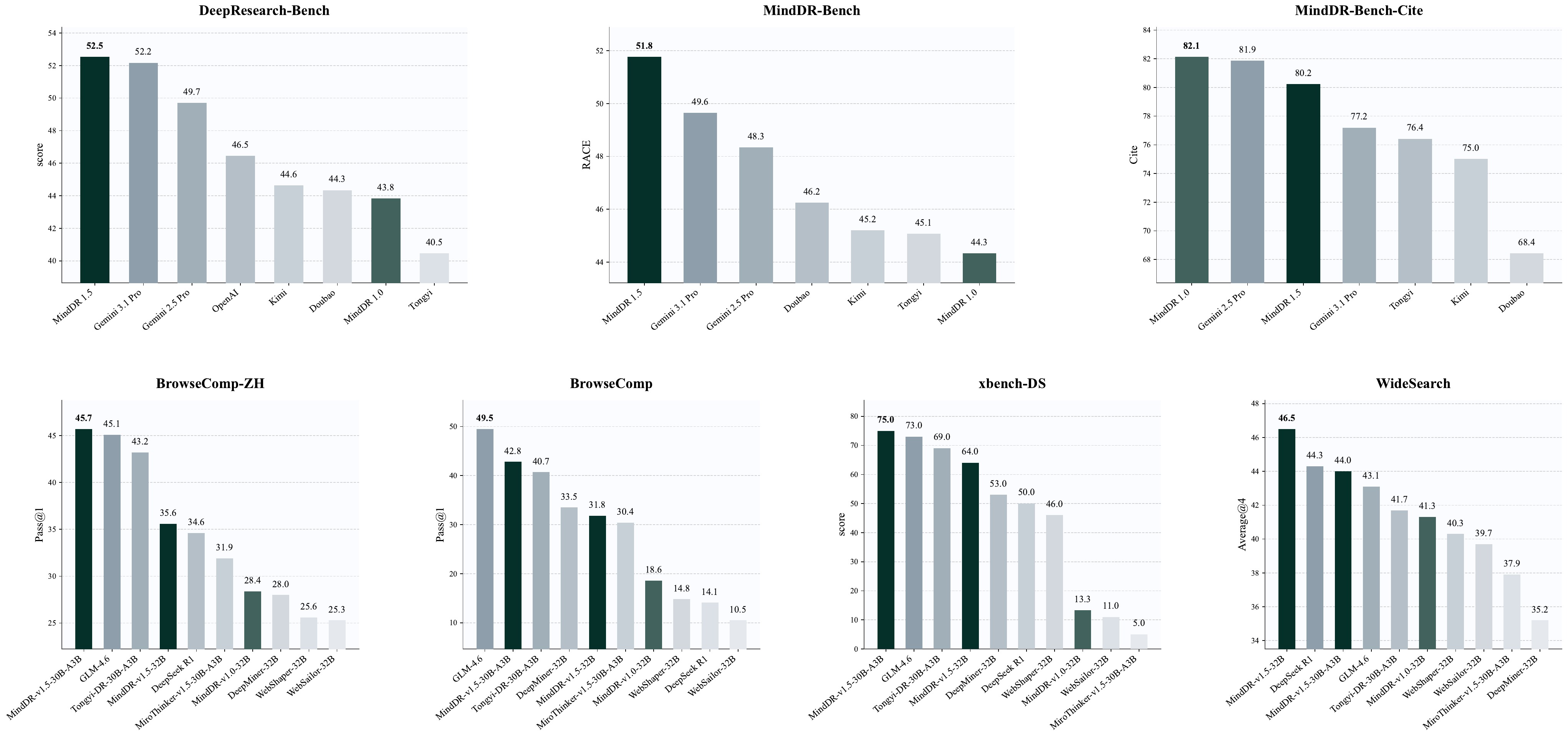}
    \caption{Benchmark Performance of MindDR, comparing with mainstream deep research products and state-of-the-art models at comparable parameter scale. MindDR 1.0 denotes the previous-generation model trained with large-scale RFT, while MindDR 1.5 denotes the model presented in this paper. MindDR 1.5 has been listed on the official DeepResearch Bench leaderboard: \href{https://huggingface.co/spaces/muset-ai/DeepResearch-Bench-Leaderboard}{https://huggingface.co/spaces/muset-ai/DeepResearch-Bench-Leaderboard}   }
    \label{fig:benchmarks}
\end{figure}


The rapid advancement of large language models (LLMs) has fundamentally transformed human workflows and daily life, substantially boosting productivity across a wide spectrum of tasks~\cite{deepseek2025r1,openai2025deepresearch}. The field has undergone a clear paradigm shift: LLMs have evolved from simple conversational chatbots to reasoners capable of complex multi-step logical inference~\cite{jason2022chain}, and further to autonomous agents that can plan, reason, and interact with external tools~\cite{shunyu2023react,timo2023toolformer,yujia2023toolllm}. Among the emerging applications of LLM agents, \emph{deep research agents} have become one of the most representative and capable product paradigms~\cite{google2024gemini2,openai2025deepresearch}.

Compared to traditional single-turn QA or retrieval-augmented generation (RAG) pipelines relying on one-shot retrieval, deep research agents support a substantially richer capability set, encompassing open-domain information retrieval, multi-source evidence verification and synthesis, long-horizon reasoning~\cite{shunyu2023tree}, external tool use and structured report generation. The paradigm gained widespread attention when Google and OpenAI introduced their powerful deep research agents ~\cite{google2024gemini2, openai2025deepresearch}, exhibiting human-level performance in domains such as scientific research and financial analysis, catalyzing rapid adoption of the deep research paradigm across academia and industry. Following these pioneering closed-source systems, significant open-source efforts have emerged to democratize deep research capabilities, including Tongyi DeepResearch~\cite{team2025tongyi,websailor2025,wu2025webdancer,team2025webweaver}, MiroThinker~\cite{team2025mirothinker} and Step-DeepResearch~ \cite{stepfun2025stepdeepresearch}. 

While the open-source ecosystem has made notable progress, deep research agents continue to face a fundamental bottleneck: the prohibitive cost of training and inference, which severely hinders practical user experience. State-of-the-art deep research systems often rely on massive foundation models (e.g., $>$100B parameters) and expensive training paradigms, such as extensive continual pre-training (mid-training), to inject domain knowledge and reasoning capabilities. Furthermore, at inference time, complex research tasks require long-horizon reasoning and multi-step tool use~\cite{shunyu2023tree}. Without explicit optimization for search efficiency, models tend to exhaust substantial computational budgets on marginally relevant exploration with limited information gain~\cite{xue2025onlinemind2web}. This inefficiency drastically increases token consumption and system latency, and also risks diluting key findings through excessive context accumulation~\cite{liang2025survey}, ultimately degrading the user experience.

Therefore, the core challenge in deep research is: \textit{how to achieve leading performance and excellent user experience using a small-sized model through low-cost training and inference?} 

To address this challenge, we present \textbf{MindDR}, a cost-effective framework that achieves state-of-the-art deep research capabilities using only \textasciitilde30B-parameter models. MindDR tackles the cost-performance trade-off by decomposing the complex research problem into specialized subtasks at the inference stage and applying a highly targeted, multi-stage training pipeline at the training stage.

\textbf{Inference-stage decomposition.} At inference time, MindDR employs a collaborative three-agent architecture---a Planning Agent, a DeepSearch Agent, and a Report Agent. This decomposition allows parallel execution of search tasks and context isolation, inherently improving inference efficiency and alleviating the burden of ultra-long context processing on a single model. The DeepSearch Agent efficiently navigates multi-step retrieval scenarios, while the Report Agent focuses on resolving information conflicts~\cite{webweaver2025} and generating human-aligned content.

\textbf{Training-stage targeted optimization.} At the training stage, rather than relying on computationally expensive mid-training or monolithic end-to-end reinforcement learning (RL), we design a four-phase training pipeline: (i)~\textit{supervised fine-tuning (SFT) for behavioral cold-start}, establishing foundational instruction-following and tool-use capabilities while maintaining low data scale; (ii)~\textit{Search-RL}, which explicitly optimizes the DeepSearch Agent's long-horizon reasoning and search efficiency through step-level credit assignment, significantly reducing redundant token consumption during inference; (iii)~\textit{Report-RL}, which employs RACE Rubrics and format-based reward shaping to specialize the Report Agent in resolving information conflicts and generating high-quality long-form content; and (iv)~\textit{preference alignment}, which further calibrates output reports to human expectations via human feedback signals.

We evaluate the performance of MindDR on deep search and deep research benchmarks such as BrowseComp(-ZH), GAIA, xbench-DS and DeepResearch Bench to verify its effectiveness. Furthermore, we also establish MindDR Bench, a curated benchmark comprising 500 deep research queries extracted from real-world user interactions with our AI assistant. Our main contributions are summarized as follows:

\begin{enumerate}
    \item \textbf{Task-driven multi-stage training pipeline for heterogeneous agents.} We design a four-stage training pipeline tailored to distinct agent requirements: SFT for behavioral cold-start and Search-RL for long-horizon reasoning and search efficiency in DeepSearch Agent, Report-RL for information conflict resolution and report quality, and preference alignment for human experience optimization in Report Agent. The Search-RL stage achieves prominent accuracy improvement on BrowseComp-ZH while reducing context and tool-call consumption compared to SFT baseline model and the subsequent Report-RL stage also improves report RACE scores.

    \item \textbf{Introduction of MindDR Bench and a Comprehensive Evaluation System.} We introduce MindDR Bench, a rigorously curated benchmark comprising 500 Chinese deep research queries extracted from real-world user interactions with Livis (Li Auto's intelligent assistant). Rather than relying on a single, abstract RACE metric for content evaluation, we propose a comprehensive evaluation system tracking the search process, assessing both the content quality and the presentation format of generated reports. MindDR Bench explicitly aligns the evaluation metrics with practical user experience, providing diverse insights for deep research community.
\end{enumerate}


The remainder of this paper is organized as follows. Section~\ref{sec:relatedworks} reviews related works on deep research agents, search reinforcement learning, and report generation reinforcement learning. Section~\ref{sec:framework} presents the MindDR multi-agent framework. Section~\ref{sec:datasyn} describes the data synthesis pipeline. Section~\ref{sec:trainingpipe} elaborates on the multi-stage training pipeline. Section~\ref{sec:results} shows main experimental results across benchmarks with comparisons with representative open-source and closed-source systems. Section~\ref{sec:conclusion} concludes the paper with a discussion of current limitations and future research directions. Section~\ref{sec:appendix} appends the prompts and examples.

\section{Related Works}
\label{sec:relatedworks}
\subsection{Deep Research Agents}
Knowledge-intensive tasks have evolved from single-turn Retrieval-Augmented Generation (RAG)~\cite{singh2025agenticrag} toward autonomous agents capable of iterative tool use and long-horizon reasoning~\cite{shunyu2023react, shunyu2023tree}. Proprietary systems like Gemini Deep Research~\cite{google2024gemini2} and OpenAI Deep Research~\cite{openai2025deepresearch} demonstrated near-human performance on complex investigative tasks spanning scientific research and financial analysis, yet their closed-source nature limits reproducibility and systematic analysis~\cite{li2026evaluating}. This opacity has spurred open-source alternatives. Tongyi DeepResearch~\cite{team2025tongyi} proposed an end-to-end optimization architecture for agentic training. MiroThinker~\cite{team2025mirothinker,mirothinker2026heavy} advanced ``interactive scaling'' by training models to handle hundreds of tool calls via reinforcement learning. WebSailor~\cite{websailor2025,li2025websailor2} focused on uncertainty-reducing web navigation, while WebWeaver~\cite{webweaver2025} explored dual-agent architectures for open-ended report generation.  Nanbeige4.1-3B~\cite{nanbeige2026} further showed that small models can be competitive through specialized training. Existing systems predominantly optimize for retrieval accuracy under a monolithic end-to-end RL objective, which is difficult to train due to the high complexity of deep research tasks.

\subsection{Search Reinforcement Learning}
Search-RL optimizes retrieval decision-making and multi-step reasoning within deep research agents, directly targeting the challenge of \textit{long-horizon reasoning} in complex information-seeking scenarios. Research in this area advances along two principal directions.

\textbf{Data Synthesis.} Bootstrapping effective policies requires high-quality retrieval-reasoning trajectories. Knowledge-graph-driven methods (e.g., WebSailor~\cite{websailor2025}, DeepDive~\cite{lu2025deepdive}) produce logically consistent data but are constrained by graph coverage and struggle in dynamic open-domain scenarios. Agent-simulation-based approaches (e.g., MiroThinker~\cite{team2025mirothinker}, Cognitive Kernel-Pro~\cite{fang2025cognitive}) improve task alignment and realism, yet incur prohibitive computational costs and lack fine-grained modeling of per-step retrieval contributions, making it difficult to provide differentiated training signals for critical versus peripheral search steps.

\textbf{Credit Assignment.} Beyond data quality, training efficiency hinges on precise credit assignment. Trajectory-level RL methods~\cite{wang2025ragen, jin2025search, dong2025arpo} broadcast a uniform reward across all steps, providing no targeted signal for critical retrievals and thus failing to explicitly optimize \textit{search efficiency}. Step-level approaches offer stronger supervision: PPO-based methods~\cite{zheng2025stepsearch} introduce per-step rewards but depend on costly critic models; branch-sampling methods (e.g., ARPO~\cite{dong2025arpo}, TreeRL~\cite{hou2025treerl}) reduce overhead yet remain infeasible for full-coverage, long-horizon tasks. MindDR addresses this gap with a lightweight step-level credit assignment mechanism that achieves fine-grained advantage estimation without critic overhead or exponential sampling complexity, enabling explicit optimization for both retrieval accuracy and search efficiency.

\subsection{Report Reinforcement Learning}
Report-RL trains models to generate well-structured, factually consistent long-form reports from retrieved evidence, directly tackling the challenges of \textit{information conflict resolution} and \textit{human-aligned report generation}. Progress has been driven by two complementary efforts.

\textbf{Alignment and Generation Frameworks.} Industrial systems apply multi-stage pipelines with structured rewards: Step-DeepResearch~\cite{stepfun2025stepdeepresearch} employs checklist-based rubrics, and Tongyi DeepResearch~\cite{team2025tongyi} develops end-to-end RL for long-text outputs. Extensive research augment this direction with reflection-guided writing (SuperWriter~\cite{wang2025superwriter}), recursive revision (Re3~\cite{yang2022re3}), and backward-inference trajectory construction (REER~\cite{wang2025reer}), collectively strengthening global planning and iterative self-correction. However, these approaches generally lack explicit mechanisms for adjudicating conflicting evidence across heterogeneous sources—a critical requirement in realistic research scenarios.

\textbf{Evaluation and Reward Design.} Classic evaluation metrics for deep research agents are the RACE rubric score comprising Comprehensiveness, Insight, Instruction Following and Readability as shown in DeepResearch Bench~\cite{du2025deepresearchbench}. Several other multi-dimensional rubric frameworks are also proposed including WritingBench~\cite{wu2025writingbench}, ResearchRubrics~\cite{sharma2025researchrubrics}, and DEER~\cite{han2025deer}, that enable reliable LLM-as-a-Judge evaluation along axes such as factual accuracy, structural coherence, and theme alignment. These rubrics in turn supply reward signals for RL algorithms including GRPO~\cite{shao2024deepseekmath}, GSPO~\cite{zheng2025gspo}, and DAPO~\cite{yu2025dapo}. Nevertheless, as indicated by FINDER~\citep{zhang2025finder}, current models still exhibit critical gaps in global logical structure and factual fidelity under ultra-long contexts, and few works incorporate human preference feedback to align outputs with real-world user expectations. MindDR targets these limitations through RACE Rubrics-based reward shaping combined with a dedicated preference alignment stage, directly optimizing report generation for both information quality and user reading experience.

\section{MindDR Framework}
\label{sec:framework}

\begin{figure}[t]
    \centering
    \includegraphics[width=1\linewidth]{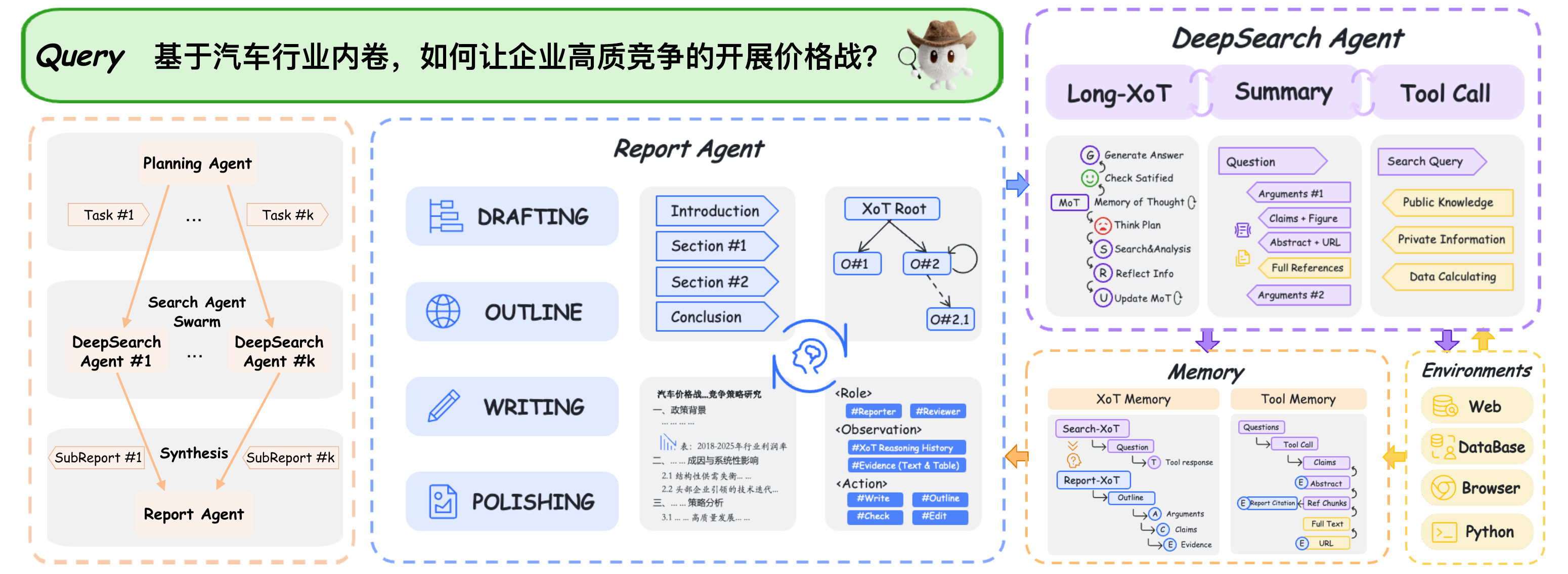}
    \caption{Overview of the MindDR multi-agent framework. A user query is first processed by the \emph{Planning Agent}, which performs intent analysis and task decomposition to produce a structured subtask specification. Each subtask is dispatched to an independent \emph{DeepSearch Agent} instance that executes a ReAct-style loop while maintaining an Extended Chain-of-Thought (XoT) reasoning trace. The resulting sub-reports are aggregated by the \emph{Report Agent}, which synthesizes a coherent, citation-grounded research report.}
    \label{fig:framework}
\end{figure}

In this section, we present the overall architecture of MindDR. As illustrated in Fig.~\ref{fig:framework}, MindDR consists of two tightly coupled components: an \emph{inference pipeline} that orchestrates multi-agent collaboration for deep research report generation (Section~\ref{sec:inference_pipeline}), and a \emph{four-phase training pipeline} that progressively builds the underlying model capabilities required by each agent (Section~\ref{sec:training_overview}). In the inference end, given a natural-language research query, the inference pipeline decomposes the problem into manageable subtasks, retrieves and synthesizes evidence from heterogeneous sources, and assembles a polished, structured report. In the training end, a four-stage pipeline equips the agents with the necessary capabilities through supervised fine-tuning, search-oriented reinforcement learning, and report-oriented reinforcement learning.


\subsection{Inference Pipeline}
\label{sec:inference_pipeline}

The inference pipeline comprises three functional agents---the Planning Agent, the DeepSearch Agent, and the Report Agent---coordinated through a shared memory substrate and a novel reasoning mechanism termed the Extended Chain-of-Thought (XoT). We describe each component below.

\paragraph{Planning Agent}When MindDR receives a user query, the Planning Agent initiates the research pipeline by analyzing user intent and decomposing the query into a set of subtasks. Then, these subtasks will be imported into the following DeepSearch Agent for parallel deep searching.

\paragraph{DeepSearch Agent}Each subtask produced by the Planning Agent is dispatched to an independent DeepSearch Agent instance, enabling parallel execution across all subtasks. Each DeepSearch Agent implements a ReAct-style agent loop~\cite{shunyu2023react}, iteratively invoking search tools to perform multi-source retrieval, evidence integration, and intermediate reasoning until the agent determines that sufficient information has been gathered to address the assigned sub-problem.



\paragraph{Report Agent}The Report Agent serves as the final synthesis stage of the inference pipeline. It receives the complete task specification and all sub-reports from the DeepSearch Agents. Based on these inputs, the Report Agent first generates a hierarchical outline, then performs global information aggregation and structural organization to produce a coherent, comprehensive, and well-structured research report in Markdown format. The Report Agent is designed to excel in several dimensions aligned with the RACE evaluation framework~\cite{du2025deepresearchbench} and realistic user experience.

\paragraph{Memory}To enable effective coordination across the multi-agent pipeline, we introduce the memory mechanism including Extended Chain-of-Thought (XoT) memory and tool memory. Unlike standard chain-of-thought prompting~\cite{jason2022chain}, which operates within a single model invocation, XoT memory extends the reasoning trace across multiple agent interactions and tool calls. DeepSearch and Report Agent maintain and append to a shared reasoning context that captures not only the current agent's thought process but also the inter-connection information between agents. Furthermore, a too-call memory module is also constructed to maintain the interactions with external environment. This shared memory enables downstream agents (e.g., the Report Agent) to access the full provenance of retrieved information, facilitating more faithful and well-grounded report generation.

\subsection{Training Pipeline Overview}
\label{sec:training_overview}
\begin{figure}
    \centering
    \includegraphics[width=1\linewidth]{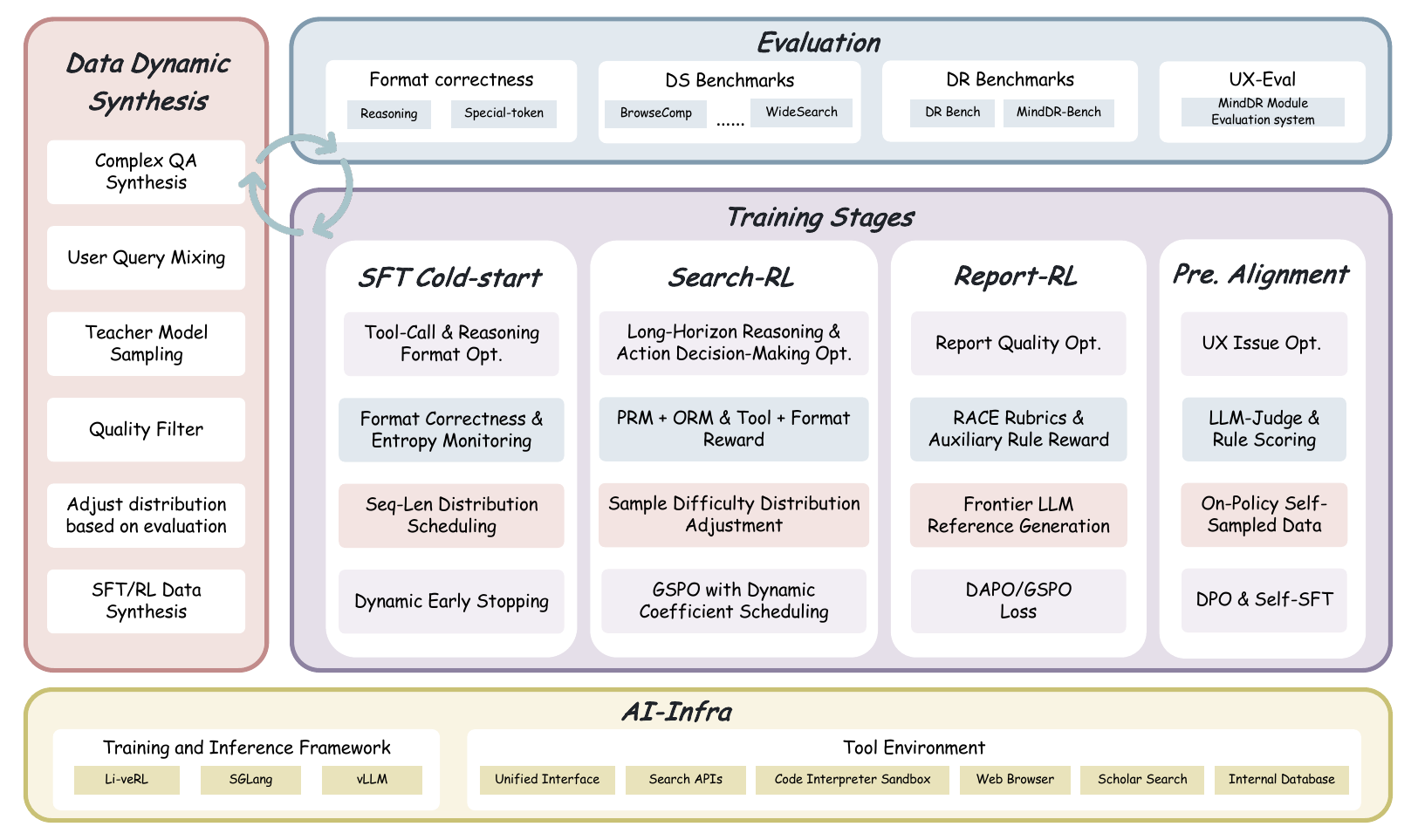}
    \caption{Four-stage training pipeline of MindDR.}
    \label{fig:training-pipe}
\end{figure}
The multi-agent MindDR system requires diverse capabilities spanning tool use, multi-step reasoning, long-form generation, and subjective quality alignment—objectives that differ substantially in their reward structure, optimization landscape, and data requirements. Rather than optimizing all objectives end-to-end, we decompose the training into a four-phase curriculum (Fig.~\ref{fig:training-pipe}), where each phase targets a well-defined capability frontier with tailored optimization algorithms and reward signals. This decomposition is guided by three principles:

\begin{itemize}[leftmargin=1.5em,itemsep=2pt]
    \item \textbf{Reward tractability.} End-to-end optimization over the full DR pipeline would require a single reward to capture tool correctness, reasoning quality, report coherence, and subjective preferences simultaneously. Such a composite reward is inevitably sparse and noisy, making credit assignment across dozens of reasoning steps intractable. Staged training decomposes this into dense, well-defined signals at each phase.
    \item \textbf{Capability dependency.} Later capabilities critically depend on earlier ones: RL exploration requires stable format adherence from SFT; report generation quality is bottlenecked by retrieval completeness from Search-RL; and preference alignment presupposes functionally correct outputs from prior phases. The ordering reflects this dependency chain.
    \item \textbf{Data efficiency.} Each phase operates on data specifically curated for its target capability, avoiding the need for expensive end-to-end trajectory annotation and enabling independent iteration on data quality per phase.
\end{itemize}

\paragraph{Phase~1: Supervised Fine-Tuning.}
SFT provides a behavioral cold-start, establishing foundational capabilities in tool invocation, ReAct-format adherence, and multi-turn reasoning patterns through behavior cloning on expert trajectories. The training extent is carefully calibrated: sufficient to ensure stable format correctness under long contexts, yet restrained to preserve policy entropy for subsequent RL exploration (Section~\ref{sec:sft}).

\paragraph{Phase~2: Search-RL.}
This phase optimizes the DeepSearch Agent's long-horizon reasoning and action decision-making ability via online RL with real tool execution. A unified GRPO/GSPO framework with dynamically scheduled rewards—progressing from tool-call correctness to process-level entity coverage to outcome-level answer accuracy—enables progressive capability acquisition without hard stage boundaries (Section~\ref{sec:search_rl}).

\paragraph{Phase~3: Report-RL.}
Report-RL targets the Report Agent's long-form generation quality. Using RACE Rubrics~\cite{du2025deepresearchbench} evaluated by LLM-as-Judge, the model is optimized along comprehensiveness, readability, insight, and instruction-following dimensions, supplemented by rule-based citation and format rewards for efficiently addressable quality issues (Section~\ref{sec:report_rl}).

\paragraph{Phase~4: Preference Alignment.}
In the generated long-form report, there exists user experience (UX) issues such as temporal correctness, table formate errors. In order to improve the report quality stably without catastrophic forgetting, MindDR adopts on-policy self-improved framework with DPO~\cite{rafailov2023dpo} and Self-SFT~\cite{chen2025selfsft} to align the final report quality with human expectations.(Section~\ref{sec:reference_align}).
\section{Data Synthesis}
\label{sec:datasyn}
\subsection{Query Synthesis}
\label{sec:query_synthesis}

\begin{figure}
    \centering
    \includegraphics[width=1\linewidth]{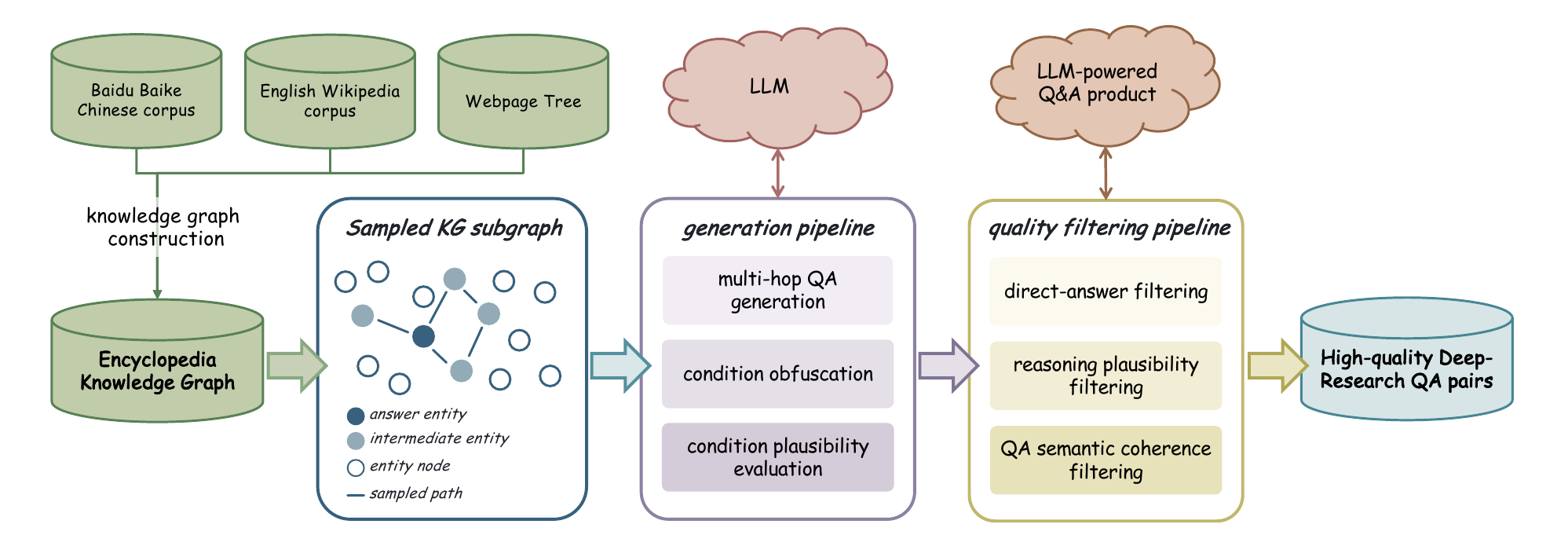}
    \caption{Overview of the knowledge-graph-grounded query synthesis pipeline, consisting of four stages: graph construction and subgraph sampling, initial QA generation, text obfuscation and complexity enhancement, and reasoning validity filtering.}
    \label{fig:dataflow}
\end{figure}

We propose an end-to-end framework for synthesizing multi-hop reasoning questions from structured knowledge graphs. The overall pipeline, illustrated in Fig.~\ref{fig:dataflow}, comprises four stages: graph construction and subgraph sampling, initial QA generation, condition obfuscation and complexity enhancement, and reasoning validity filtering.

\textbf{Graph Construction and Subgraph Sampling.} We construct unified knowledge graphs from Baidu Baike and English Wikipedia corpora, organizing information through entity-attribute relationships and sample nodes and paths over web-page tree structures. Beyond connectivity, subgraph sampling enforces three key constraints: each hop in the reasoning path must correspond to retrievable or verifiable conditions in a real search environment (\emph{reasoning reachability}); direct associations between intermediate nodes and answers are constrained to eliminate shortcuts, ensuring every hop contributes necessary information to the reasoning chain (\emph{path necessity}); and reasoning paths in multi-branch subgraphs maintain semantic and structural independence, preventing implicit cross-branch associations from reducing combinatorial complexity (\emph{structural independence}). Together, these constraints guaranty that subgraphs are inferable, non-simplifiable, and verifiable.

\textbf{Multi-hop QA Generation.} Given a subgraph structure, we prompt a state-of-the-art LLM to transform explicit structured relations into implicit natural-language QA pairs. Generation constraints ensure questions comprehensively cover key subgraph information while avoiding direct exposure of intermediate reasoning nodes, tightly binding the solution process to the graph structure and yielding high-quality multi-hop QA data for deep-search applications.

\textbf{Condition Obfuscation and Plausibility
Evaluation.} To increase question difficulty while preserving solvability, we apply controlled obfuscation to the initial QA pairs: low-value attributes are combined with other conditions or replaced with more distinctive descriptions to ensure condition retrievability; overly explicit clues are weakened or rephrased to force models to synthesize multiple conditions, preventing any single strong constraint from dominating the solution; obfuscated expressions must remain semantically consistent with the original, preferring natural, human-cognition-aligned formulations; and condition combinations are continuously monitored for their effect on the solution space, with constraints supplemented or restructured as needed to maintain answer uniqueness.

\textbf{Quality Filtering.} Generated questions are passed through a multi-stage data quality filtering pipeline: a direct-answer test first removes questions solvable without multi-step reasoning; condition plausibility evaluation and reasoning plausibility filtering then eliminate samples with logical contradictions or insufficient real-retrieval support; finally, QA semantic coherence filtering performs a holistic quality assessment combining automated scoring and human auditing, ensuring only high-quality samples enter the final training set.

\subsection{Query Source Diversification and Mixing}
\label{sec:query_mix}

The query synthesis pipeline (Section~\ref{sec:query_synthesis}) produces structured queries equipped with complete multi-hop reasoning chains, ground-truth answers, and intermediate reasoning steps. Grounded in high-quality encyclopedic corpora with well-defined inference paths, these queries are well-suited for constructing structured deep reports and serve as the primary training signal across all pipeline stages.

However, the sampling space of knowledge-graph queries is inherently constrained by corpus coverage: queries involving rare entities exhibit distributional shift relative to high-frequency user demands, and capabilities acquired from such data may not transfer reliably to real-world usage patterns. To mitigate this gap, we supplement synthesized queries with real user queries mined from online interaction logs, yielding a complementary set that faithfully reflects genuine user intent. Across all training stages, knowledge-graph-synthesized queries and real user queries are mixed at carefully calibrated proportions to balance controllability with ecological validity.

Beyond training, a representative subset of these real user queries is further curated into a dedicated evaluation benchmark—\textbf{MindDR Bench}—which we describe next.

\subsection{MindDR Bench}
\label{sec:mindr_bench}

While the diversified query mixture described above strengthens training, a parallel challenge arises on the evaluation side. During the course of benchmark investigation, we identify two primary limitations in existing deep-research benchmarks that restrict their utility for guiding practical system development. First, there remains a notable scarcity of authentic Chinese queries; the majority of existing test cases are synthetic or derived from academic datasets, introducing distributional bias relative to actual user experience. Second, prevailing evaluation frameworks rely predominantly on macro-level metrics (\emph{e.g.}, a single aggregate RACE score). Given the inherently long execution horizons of deep research tasks, such coarse-grained metrics fail to provide actionable, fine-grained feedback to intermediate pipeline modules, thereby hindering targeted iteration and continuous improvement.

To bridge this gap, we introduce \textbf{MindDR Bench}, a rigorous benchmark explicitly designed to reflect genuine user intent and support comprehensive, actionable evaluation.

\textbf{Query Curation.}\quad We construct MindDR Bench by mining 500 deep research queries directly from the online interaction logs of real users with Li Auto's intelligent assistant. Quality and complexity are ensured through a two-stage hybrid filtering pipeline: a state-of-the-art LLM first prescreens candidates for the required reasoning depth, followed by expert annotation and review. The resulting queries span 16 distinct domains—including automotive, travel, technology, and finance—faithfully capturing authentic, high-complexity research demands grounded in automotive industry scenarios.

\textbf{Comprehensive Evaluation System.}\quad To overcome the limitations of macro-level scoring, we propose a fine-grained, multi-dimensional \textbf{MindDR Module Evaluation} system built upon the foundational DeepResearch Bench framework~\cite{du2025deepresearchbench}. Rather than relying solely on a holistic RACE score, our system systematically decomposes evaluation across four critical stages of the deep research pipeline, as detailed in Table~\ref{tab:module_evaluation}.

\begin{table}[htbp]
    \centering
    \caption{MindDR Module Evaluation system across four critical pipeline stages.}
    \label{tab:module_evaluation}
    \small
    \renewcommand{\arraystretch}{1.25}
    \begin{tabular}{@{}p{3.2cm}p{4.0cm}p{5.8cm}@{}}
    \toprule
    \textbf{Evaluation Module} & \textbf{Evaluation Metrics} & \textbf{Evaluation Items} \\
    \midrule
    \multirow{2}{3.2cm}{\textbf{Reasoning Trajectory}} & \multirow{2}{4.0cm}{Thinking Efficiency}
        & 1.\ Reflection turns count \\
        & & 2.\ Search query repetition rate \\
    \midrule
    \multirow{2}{3.2cm}{\textbf{Tool Call}} & \multirow{2}{4.0cm}{Correctness of tool use}
        & 1.\ Proportion of usage for each tool \\
        & & 2.\ Tool call failure rate \\
    \midrule
    \multirow{2}{3.2cm}{\textbf{Outline Generation}} & \multirow{2}{4.0cm}{Correctness of outline logic}
        & 1.\ Outline title miss rate \\
        & & 2.\ Incorrect directory hierarchy count \\
    \midrule
    \multirow{2}{3.2cm}{\textbf{Report Generation}} & \multirow{2}{4.0cm}{Correctness of content logic}
        & 1.\ Tense error rate \\
        & & 2.\ Valid format tables rate \\
    \bottomrule
    \end{tabular}
\end{table}
By combining these modular process indicators with content-focused RACE metrics, our evaluation system comprehensively captures both the intermediate behaviors and the final presentation of the system. This granular feedback mechanism explicitly aligns model performance with practical user experience metrics and allows for targeted iteration of individual pipeline modules, significantly accelerating the iterative development of the MindDR pipeline.

\subsection{SFT Data Synthesis}
\label{sec:sft_data_synthesis}

SFT data endows the model with cold-start capabilities spanning tool invocation, structured formatting, and multi-turn reasoning. We build an end-to-end synthesis system covering multi-source sampling, multi-tier filtering, standardized post-processing, and automated configuration. All data follow the ReAct paradigm, decomposing complete reasoning trajectories into independent steps organized as multi-turn dialogs; each step contains a thought $T_t$, an action $A_t$, and an observation $O_t$, enabling unified decision-process learning. The resulting trajectory corpus serves a dual purpose: one partition is used directly for supervised fine-tuning, while the other is repurposed as seed data for Report-RL training (Section~\ref{sec:report_rl_data_synthesis}).

\textbf{Multi-Source Sampling and Scale.}\quad We construct approximately 12K high-quality trajectories from three complementary sources: (i)~knowledge-graph trajectories (60\%), derived from the multi-hop reasoning data described in Section~\ref{sec:query_synthesis} and spanning 1--5 hops at varying complexity; (ii)~real-world scenario trajectories (35\%), covering automotive, technology, transportation, and industrial domains to ensure ecological validity; and (iii)~human-annotated high-difficulty edge cases (5\%) for robustness. All trajectories are generated via multi-model parallel sampling in a simulation environment closely aligned with the online inference stack, ensuring sufficient data diversity.

\textbf{Trajectory Diversity Design.}\quad Data distribution is controlled along three dimensions: task complexity (easy: 1--2 hops, 40\%; medium: 3 hops, 35\%; hard: 4--5 hops, 25\%), trajectory length (5--30 steps, avoiding fixed-length decision biases), and tool invocation patterns (sequential retrieval, parallel verification, and hierarchical deepening).

\textbf{Long-Context Data Augmentation.}\quad We adopt a progressive length-generalization strategy proceeding through a base phase (average 8K tokens), an extension phase (32K/64K, 30\% of the mixture), and an extreme-length phase (128K, 15\%). Positional encoding is fine-tuned by resampling trajectories that contain redundant observations such as history reviews and intermediate summaries. This strategy raises 128K-context format correctness from 72\% to 94\%.

\textbf{Multi-Tier Quality Filtering and Configuration.}\quad We establish a three-tier filtering system combining rule engines and LLM-based evaluation. The \emph{pre-admission} stage filters out simple queries, retaining only complex open-ended questions that require multi-hop reasoning. The \emph{process-validation} stage performs real-time verification of reasoning logic, tool invocation accuracy, and content relevance. The \emph{final-approval} stage conducts multi-dimensional quality assessment—covering format compliance, factuality, instruction following, and logical coherence—supplemented by human auditing. After uniform format validation, metadata completion, and multi-dimensional tagging via automated post-processing, a dynamic configuration strategy prioritizes complex multi-turn reasoning samples (60--70\%) while maintaining balanced coverage of tool invocation and report generation capabilities.

\subsection{Search-RL Data Synthesis}
\label{sec:searchrl_data_synthesis}

Search-RL training data is built on the multi-hop reasoning queries generated in Section~\ref{sec:query_synthesis}. The key design principle is to preserve the complete reasoning chain from question to answer during synthesis, providing the supervision signals required for step-level process reward modeling (PRM) and outcome reward modeling (ORM). We use approximately 35K synthesized queries as the training data foundation.

\textbf{Entity Annotation and Reward Data Construction.} During query generation, we extract and retain the intermediate entity nodes, relational transition paths, and reasoning dependency structure from the knowledge graph, forming the key entity set $\mathcal{E} = \{e_1, \ldots, e_M\}$ for each query. This entity set is used directly for string-matching-based PRM verification during training (see Section~\ref{sec:search_rl}), requiring no additional LLM inference and substantially reducing reward computation cost. The ground-truth answers required for ORM are recorded at query generation time, supporting final-output matching verification.

\textbf{Difficulty Annotation and Distribution Control.} Each query is annotated with a three-level difficulty label based on reasoning hop count (1--5 hops), intermediate entity category complexity, and retrieval difficulty (entity ambiguity, information sparsity). These labels support dynamic sampling during training, where the actual sampling proportions are adjusted adaptively according to ORM accuracy on the validation set (see Section~\ref{sec:search_rl}); the annotation layer is responsible only for providing standardized difficulty metadata, without prescribing fixed curriculum learning ratios.

\textbf{Data Format.} Each training instance contains a query, the intermediate entity set $\mathcal{E}$, the ground-truth answer, and a difficulty label. The format is consistent with the SFT data described in Section~\ref{sec:sft_data_synthesis}, using the ReAct multi-turn dialogue structure, and can be ingested directly by the RL training pipeline.

\subsection{Report-RL Data Synthesis}
\label{sec:report_rl_data_synthesis}

Report-RL training requires query–report pairs augmented with fine-grained scoring rubrics. Rather than constructing an independent data pipeline, we reuse the high-quality trajectory corpus produced during SFT data synthesis (Section~\ref{sec:sft_data_synthesis}) as the foundation, and derive two complementary data formats—long-form and short-form—to balance training fidelity with data efficiency.

\paragraph{Long-form Data Synthesis.}
Each long-form training instance comprises six components: a query, a system prompt, upstream deep-search retrieval data, an outline, RACE Rubrics, and a reference report. The query, system prompt, retrieval data, outline, and reference report are jointly drawn from the SFT trajectory corpus, preserving the original retrieval context and reasoning chain. The RACE Rubrics are synthesized separately: queries from the SFT set, together with the RACE Rubrics generation template from DeepResearch Bench~\cite{du2025deepresearchbench}, are fed into a strong LLM to produce query-specific scoring criteria tailored to each individual query. These rubrics subsequently serve as part of the reward model's input prompt during training, enabling differentiated evaluation of reports generated for different queries. An illustrative example of the RACE Rubrics is provided in Appendix~\ref{sec:race_rubrics_example}.

\paragraph{Short-form Data Synthesis.}
\label{sec:short_form_data}
Long-form instances that carry full upstream retrieval results are expensive to sample and inevitably introduce retrieval noise. We therefore introduce a complementary short-form synthesis strategy that reduces data collection cost while substantially expanding the volume of usable training data. Each short-form instance consists of a query, a system prompt, RACE Rubrics, and a reference report—deliberately omitting the upstream retrieval content. Queries are again sourced from the SFT trajectory corpus, and the RACE Rubrics are directly reused from the long-form pipeline to maintain evaluation consistency. The reference reports, however, are \emph{newly synthesized}: a strong LLM is prompted with both the query and the dimension-specific evaluation criteria from the Rubrics, guiding it to produce a high-quality report that explicitly addresses each RACE dimension. By decoupling report generation from the retrieval stage, this strategy yields cleaner supervision signals while preserving alignment with the rubric-based reward framework. The synthesis prompts are provided in Appendix~\ref{sec:short_text_prompt}.

The overall data synthesis process for both formats is illustrated in Fig.~\ref{fig:report_rl_pipeline}.

\section{Training Pipeline}
\label{sec:trainingpipe}
The DR system involves diverse task types and multi-level capability demands. We design a four-stage training pipeline: SFT cold-start training establishes instruction-following and basic tool invocation; Search-RL builds long-horizon reasoning and complex search capabilities through online reinforcement learning; Report-RL specializes the model in producing high-quality long-form reports; and human preference alignment closes the gap between RL-optimized behavior and nuanced user expectations.

\subsection{Supervised Fine-Tuning}
\label{sec:sft}
\paragraph{Training Objective and Data Representation.}
SFT provides cold-start capabilities for tool invocation, format adherence, and multi-turn reasoning patterns, establishing a policy foundation for subsequent RL. As described in Section~\ref{sec:sft_data_synthesis}), training data follows the ReAct paradigm with trajectories decomposed into independent steps. At each step $t$, the model predicts thought content $T_t$ and tool invocation action $A_t$ based on history $H_{<t}$. Tools are not executed during training; observations $O_t$ are pre-recorded as contextual inputs.

Given dataset $(x, H) \sim \mathcal{D}_{\text{SFT}}$, the training objective is the standard autoregressive language modeling loss:

\begin{equation}
\mathcal{L}_{\text{SFT}}(\theta) = -\mathbb{E}_{(x,H)} \left[ \sum_{t=1}^{T_H} \log \pi_\theta (T_t, A_t \mid x, H_{<t}) \right]
\end{equation}

This objective performs behavior cloning on expert trajectories, theoretically optimizing a lower bound of the RL policy gradient objective, thereby providing a reasonable initial policy distribution and avoiding the prohibitive sample complexity of training from a random policy.

\paragraph{Training Strategy and Long-Context Enhancement.}
We curate approximately 15K trajectories through sampling and filtering (composition detailed in Section~\ref{sec:sft_data_synthesis}), with a systematic adjustment to the data distribution: earlier training is dominated by shorter, simpler samples with an average context length of around 8K tokens, establishing stable format adherence and basic tool invocation; as training progresses, longer and harder samples are gradually introduced, with 32K--64K context data comprising approximately 30\% of the mixture and 128K data approximately 15\%, alongside length-adaptive positional encoding fine-tuning. This curriculum data arrangement improves 128K-context format correctness from 72\% to 94\% while maintaining short-context performance.

\paragraph{The Delicate Balance Between SFT and RL.}
SFT training extent has a non-intuitive trade-off with subsequent RL effectiveness, directly determining the reasonable boundaries of the data volume and distribution described above. Under-trained models (20--40K samples) retain good exploration capacity but exhibit insufficient format adherence under long contexts, yielding fewer than 30\% valid trajectories in early RL. Over-trained models (beyond 200K samples) achieve near-perfect format correctness but exhibit rigidity during RL sampling---over-fitting to the surface patterns of expert trajectories causes sampled rollouts to be highly similar, entropy to collapse rapidly, and gradient signals to vanish. Theoretically, SFT optimizes a lower bound of the RL policy gradient objective: excessive training compresses the policy's support set, causing $\pi_\theta(\tau)$ to assign negligibly low probability to non-expert trajectories even when they may yield higher rewards; simultaneously, an over-fitted reference policy $\pi_{\text{ref}}$ causes KL divergence $D_{\text{KL}}(\pi_\theta \| \pi_{\text{ref}})$ to increase rapidly in early RL, triggering KL penalties that suppress exploration.

\paragraph{Training Termination Criteria.}
Based on the above analysis, we adopt \textbf{long-context format correctness} as the core early stopping metric to quantitatively determine the appropriate SFT termination point. In RL training, each query requires $G=8$ sampled trajectories to compute group relative advantages (GRPO), with at least 2 valid trajectories required for stable advantage estimation. By the binomial distribution, format error rate $p$ must satisfy the probability of at least 2 successes $\geq 0.95$, yielding $p \leq 0.0253$ (i.e., error rate below 2.53\%). Accordingly, every 10K steps we measure format error rates at 64K and 128K context lengths on a validation set; early stopping is triggered when both fall below 2.5\% and training loss plateaus. We additionally require that policy entropy at this point be no lower than 90\% of its mid-training value (at 60K steps), ensuring sufficient exploration capacity is retained. In practice, these conditions are typically satisfied at 100K--150K high-quality samples, achieving an optimal balance between competence and plasticity.

\begin{figure}[t]
    \centering
    \includegraphics[width=0.95\linewidth]{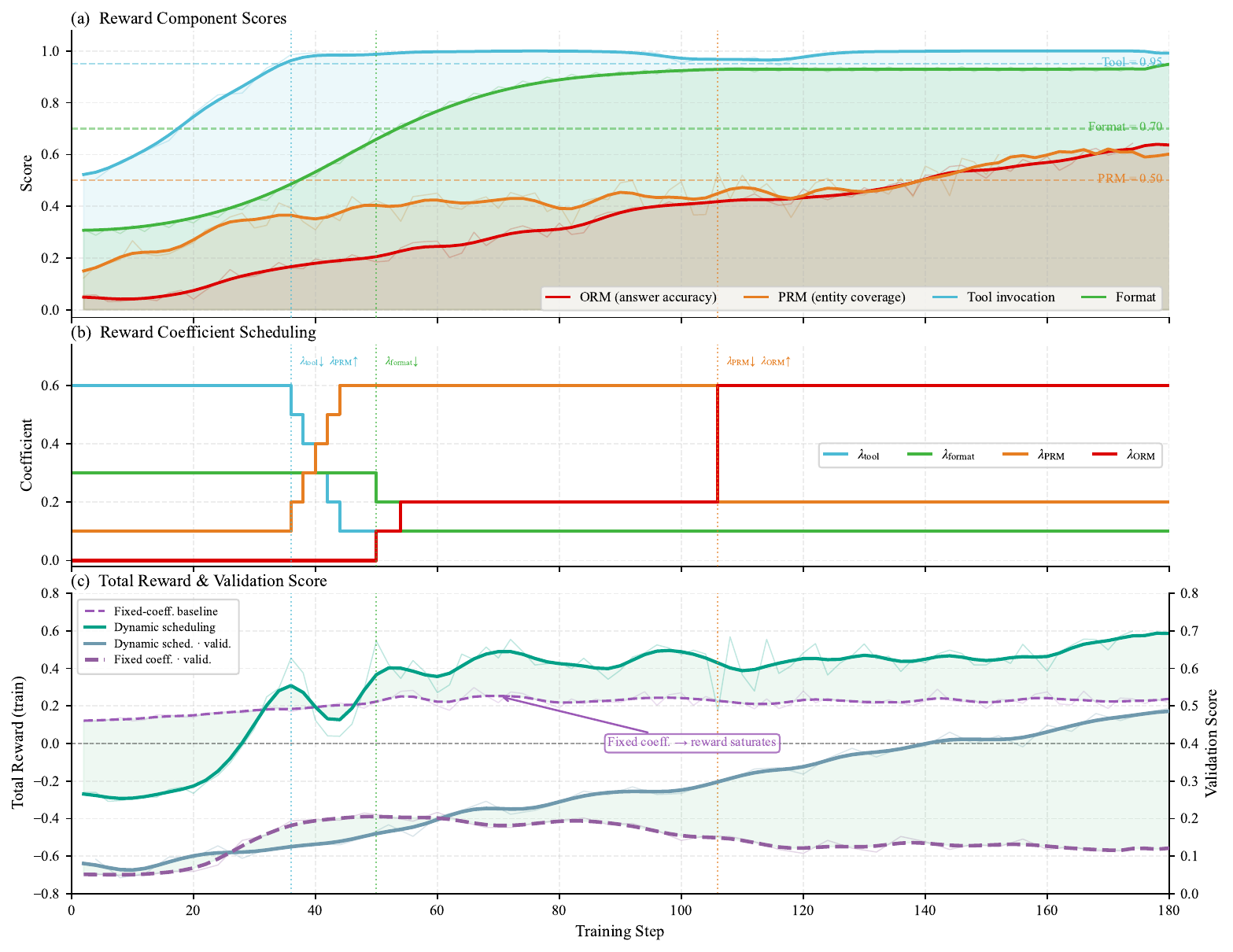}
    \caption{Training dynamics of Search-RL over 180 steps.
    \textbf{(a)} Reward component scores: ORM (answer accuracy), PRM (average entity coverage), tool invocation success, and format compliance; dashed horizontal lines mark the scheduling thresholds and dotted vertical lines indicate the three transition events.
    \textbf{(b)} Reward coefficient scheduling: $\lambda_{\text{tool}}$, $\lambda_{\text{format}}$, $\lambda_{\text{PRM}}$, and $\lambda_{\text{ORM}}$ evolve through three threshold-triggered phases as each capability saturates.
    \textbf{(c)} Total reward curve, annotated with the three key coefficient adjustment events.}
    \label{fig:reward_schedule}
\end{figure}

\subsection{Search Reinforcement Learning (Search-RL)}
\label{sec:search_rl}
The Search-RL stage targets three core challenges of deep search agents: accurate tool invocation, analytical reflection over intermediate reasoning, and information consistency with reasoning correctness in long-context complex tasks. We adopt a unified GRPO-based RL framework that progressively builds capabilities from basic tool invocation to complex long-horizon reasoning within a single continuous training process, through dynamic scheduling of reward weights and training data difficulty. This avoids the hyperparameter sensitivity and capability degradation risks associated with hard-staged decomposition.

\paragraph{Environment.}
We build a large-scale training and inference environment based on a targeted optimization of the veRL framework.
Li-veRL extends the native veRL framework along two dimensions: expert routing optimization and expert load balancing for improved MoE training stability; and efficiency enhancements including asynchronous trajectory generation, inter-trajectory asynchronous execution, and asynchronous tool calling, achieving a 2.9$\times$ speedup over native veRL. Training and inference stages employ fully consistent sampling and execution flows, fundamentally eliminating distribution shift.
Drawing on the engineering experience of Tongyi DeepResearch~\cite{team2025tongyi}, we prioritize stability at the tool layer. All tool calls are routed through a unified entry layer responsible for traffic control, exception retry, and result caching, ensuring consistent tool behavior, reducing overall tool error rate to below 0.1\%, and providing clear error feedback to the model to prevent repetitive failure patterns.

\textbf{Tool Capabilities.} The model has access to three categories of search capabilities: internal knowledge search, external web search, and academic literature search. Internal knowledge search uses a proprietary search engine combined with a tens-of-billions-scale high-quality internal knowledge base, providing superior coverage in company-relevant domains (automotive, technology, finance) while substantially reducing tool invocation cost. External search tools automatically route queries to the most suitable search providers (including Sogou, Bing, and Quark) based on query content and category tags, ensuring optimal search coverage and quality. The system additionally provides web crawling and document processing tools for full-text retrieval to support deep reading and information extraction.

For each sampled query $x$, the model generates a trajectory $H = \{(T_t, A_t, O_t)\}_{t=1}^{T}$ within the above environment, where $T_t$ is the reasoning content at step $t$, $A_t$ the model action (thought or tool call), and $O_t$ the environment-returned observation. Steps are appended sequentially to the context until a final answer is produced.

\paragraph{Sampling and Optimization Objective.}
The base optimization framework is Group Relative Policy Optimization (GRPO): for each input $x$, we sample $G$ trajectories $\{H_1, \dots, H_G\}$ and compute group-relative advantages:
\begin{equation}
\hat{A}(x, H_i) = R(x, H_i) - \frac{1}{G} \sum_{j=1}^{G} R(x, H_j)
\end{equation}
The policy optimization objective maximizes expected advantages while constraining policy drift via a KL penalty:
\begin{equation}
\mathcal{L}_{\text{GRPO}}(\theta) = \mathbb{E}_{x \sim \mathcal{D},\, H \sim \pi_\theta} \left[ \hat{A}(x, H) \cdot \log \pi_\theta(H|x) - \beta\, D_{\text{KL}}\!\left(\pi_\theta(\cdot|x) \| \pi_{\text{ref}}(\cdot|x)\right) \right]
\end{equation}
where $\pi_{\text{ref}}$ is the reference policy (initialized from SFT) and $\beta$ controls KL constraint strength.

GRPO yields stable and effective training on dense models. However, the sparse activation property of MoE models exacerbates instability: after one or more gradient updates, the expert networks activated by the same response may shift substantially, causing the activation paths of trajectories sampled under the old policy to be inconsistent with those of the current policy, thereby violating the importance sampling assumption and introducing large gradient estimation bias. We attempted a ``Routing Replay'' technique---forcing the target policy to activate the same experts as the old policy during updates---to mitigate activation path drift, but experiments showed that training remained unstable. Consequently, for MoE Search-RL training, we adopt GSPO~\cite{zheng2025gspo} (Group Sequence Policy Optimization).

The key innovation of GSPO is elevating the importance ratio from the token level to the sequence level with length normalization, unifying the numerical range across responses of varying lengths and reducing variance. Let $\{y_i\}_{i=1}^G$ be the response group sampled from the old policy $\pi_{\theta_{\text{old}}}$ for query $x$, and $\hat{A}_i$ the group-relative advantages. The GSPO objective is:
\begin{equation}
J_{\text{GSPO}}(\theta) = \mathbb{E}_{x \sim \mathcal{D},\, \{y_i\} \sim \pi_{\theta_{\text{old}}}} \left[ \frac{1}{G} \sum_{i=1}^{G} \min\!\left( s_i(\theta)\,\hat{A}_i,\;\operatorname{clip}(s_i(\theta), 1{-}\varepsilon, 1{+}\varepsilon)\,\hat{A}_i \right) \right]
\end{equation}
where the sequence-level importance ratio $s_i(\theta)$ is defined as the exponentiated mean of token-level log-probability ratios:
\begin{equation}
s_i(\theta) = \left(\frac{\pi_\theta(y_i \mid x)}{\pi_{\theta_{\text{old}}}(y_i \mid x)}\right)^{\!\frac{1}{|y_i|}} = \exp\!\left(\frac{1}{|y_i|} \sum_{t=1}^{|y_i|} \log \frac{\pi_\theta(y_{i,t} \mid x, y_{i,<t})}{\pi_{\theta_{\text{old}}}(y_{i,t} \mid x, y_{i,<t})}\right)
\end{equation}
Compared to GRPO's direct cumulative product of token-level probability ratios, the sequence-level $s_i(\theta)$ naturally averages over local expert routing changes, effectively suppressing ratio fluctuations induced by MoE sparse activation, enabling the clip constraint to function stably and preventing search capability degradation.

\paragraph{Dynamic Reward.}

The reward function comprises four signal types, all evaluated via LLM-as-Judge: each assessment involves 3 independent models producing binary (0/1) judgments with rationale, aggregated via majority vote. Both ORM and PRM employ this mechanism to improve evaluation consistency and robustness.

\textbf{Step-level Reward Definitions.} Tool invocation reward and format reward are computed independently at each trajectory step $t$. Let $c_t \in \{0,1\}$ denote the success indicator for the tool call at step $t$; the tool invocation reward is:
\begin{equation}
r_{\text{tool}}(t) =
\begin{cases}
+0.1 & c_t = 1 \\
-0.2 & c_t = 0 \text{ and } c_{t-1} = 0 \quad \text{(consecutive failure)} \\
-0.1 & c_t = 0 \text{ and } c_{t-1} = 1 \quad \text{(isolated failure)}
\end{cases}
\end{equation}
The format reward checks the structural validity of each step's output:
\begin{equation}
r_{\text{format}}(t) =
\begin{cases}
+0.1 & \text{output format correct} \\
-0.2 & \text{output format incorrect}
\end{cases}
\end{equation}
Empirically, applying stronger negative penalties (rather than positive incentives) for error-prone behaviors such as format and tool invocation failures leads to faster error avoidance.

\textbf{PRM.} The process reward is evaluated against the set of key intermediate entities $\mathcal{E} = \{e_1, e_2, \ldots, e_M\}$ associated with the query. Entities are verified via string matching within the reasoning content $T_t$ and tool call observations $O_t$ at each step. If entity $e_j$ is detected at any step, that step receives the corresponding entity score contribution. The final PRM reward is the ratio of cumulatively observed entities to the total entity count:
\begin{equation}
R_{\text{PRM}}(H) = \frac{1}{M} \sum_{j=1}^{M} \hat{e}_j, \qquad \hat{e}_j = \mathbb{1}\!\left[\exists\, t \in [1,T]: e_j \in T_t \cup O_t\right]
\end{equation}
This design requires no LLM inference for verification, is low-cost and scalable, and provides stable intermediate process supervision signals.

\textbf{ORM.} ORM evaluates the overall correctness of the final answer via the same LLM-as-Judge majority-vote mechanism, yielding $R_{\text{ORM}}(H) \in \{0, 1\}$.

\textbf{Trajectory-level Composite Reward.} The total reward for trajectory $H$ is defined as:
\begin{equation}
R(x, H) = \lambda_{\text{ORM}}\, R_{\text{ORM}}(H) + \lambda_{\text{PRM}}\, R_{\text{PRM}}(H) + \frac{1}{T}\sum_{t=1}^{T}\!\left[\lambda_{\text{tool}}\, r_{\text{tool}}(t) + \lambda_{\text{format}}\, r_{\text{format}}(t)\right]
\end{equation}
All coefficients satisfy the normalization constraint $\lambda_{\text{ORM}} + \lambda_{\text{PRM}} + \lambda_{\text{tool}} + \lambda_{\text{format}} = 1$, where $\lambda_{\text{ORM}} = 1 - \lambda_{\text{tool}} - \lambda_{\text{format}} - \lambda_{\text{PRM}}$ is an implicit variable that adjusts automatically as the other three are tuned, constrained to $[0,\, 0.5]$. 

The four coefficients follow a threshold-triggered scheduling strategy, initialized at $(\lambda_{\text{tool}}, \lambda_{\text{format}}, \lambda_{\text{PRM}}, \lambda_{\text{ORM}}) = (0.6, 0.3, 0.1, 0.0)$ and adjusted across three phases: as tool invocation saturates, $\lambda_{\text{tool}}$ is reduced and the released weight transferred to $\lambda_{\text{PRM}}$; once format compliance stabilizes, $\lambda_{\text{format}}$ is similarly reduced; and when PRM coverage matures, dominance shifts to $\lambda_{\text{ORM}}$. This adaptive coupling ensures each capability receives focused supervision at its appropriate training stage, mirroring a ``grokking'' progression from basic skills to deep reasoning (Fig.~\ref{fig:reward_schedule}).

\paragraph{Dynamic Data.}
To complement the reward scheduling, we dynamically manage training data difficulty. At fixed training step intervals, we sample and evaluate the current policy on a validation set, compute ORM accuracy across difficulty bins, and adjust the sampling proportions of different difficulty levels in the next training batch accordingly---targeting an ORM accuracy of 10\%--50\% per sampling round. An accuracy below 10\% indicates tasks are too hard, yielding overly sparse rewards and ineffective gradients; above 50\% indicates tasks are too easy, with policy saturation and diminishing returns. By maintaining the model within this ``effective learning zone'', dynamic data scheduling works in concert with dynamic reward scheduling to sustain sufficient and effective learning signals throughout training.

\begin{figure}[t]
    \centering
    \includegraphics[width=0.95\linewidth]{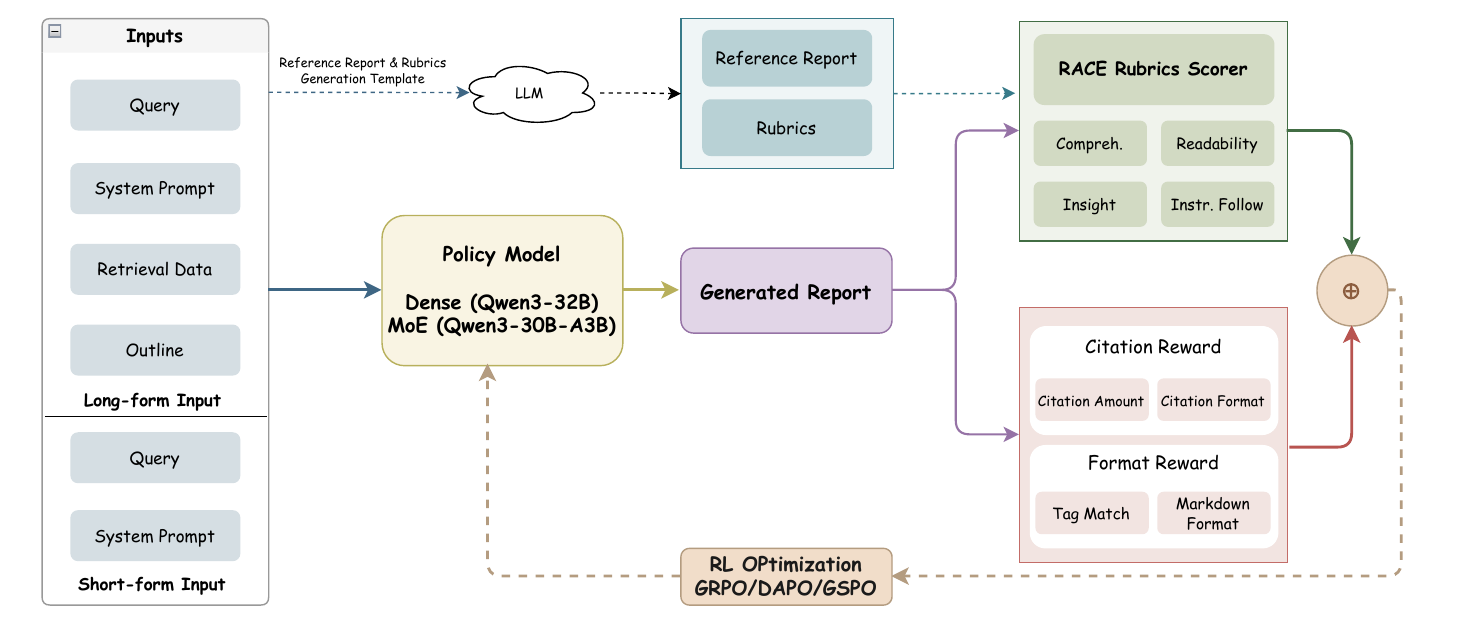}
    \caption{Overview of the Report-RL framework. Given a long-form input, the policy model and a frontier LLM such as Gemini 3.1 Pro generate sample report and reference report, respectively. The frontier LLM is also used to generate RACE Rubrics on which the sample report is evaluated. MindDR surpasses the performance of the distilled frontier LLM on DeepResearch Bench and MindDR Bench as shown in Fig.~\ref{fig:benchmarks} . }
    \label{fig:report_rl_pipeline}
\end{figure}

\subsection{Report Reinforcement Learning}
\label{sec:report_rl}

Report-RL is a reinforcement learning framework specifically designed for long-form report generation. The core idea is to leverage LLM-based evaluators scoring model outputs against structured RACE Rubrics to form reward signals for RL training. This stage focuses on improving the model's ability to produce comprehensive, readable, insightful, and instruction-following reports.

\paragraph{Framework and Environment.}

The overall Report-RL pipeline is illustrated in Fig.~\ref{fig:report_rl_pipeline}.
The training framework adopts the same Li-veRL unified infrastructure as Search-RL, supporting asynchronous inference generation and asynchronous reward computation to improve training throughput for long-form generation. Unlike Search-RL, Report-RL involves neither complex tool invocations nor multi-turn environment interactions, requiring no additional tool layer or external service dependencies; environment feedback is provided entirely by the scoring model, keeping the setup simple and deployment cost low.

Given an input consisting of a query, system prompt, upstream deep search retrieval data, and an outline, the base model generates a report. The generated report, together with pre-computed RACE Rubrics, is evaluated by a scoring model along four dimensions---\textbf{comprehensiveness}, \textbf{readability}, \textbf{insight}, and \textbf{instruction following}---combined into a weighted composite reward signal. The Rubrics design follows the RACE evaluation framework from DeepResearch Bench~\cite{du2025deepresearchbench}; the scoring prompt is detailed in Appendix~\ref{sec:scoring_prompt}.

\paragraph{Reward Design.}
In RL training practice, auxiliary reward signals are commonly introduced alongside a core reward to address specific quality issues. During our training we observe systematic errors such as tense inconsistency and malformed table output. Our guiding principle is to proactively categorize problems exposed during training: those \emph{detectable by rules without LLM inference} are incorporated as auxiliary RL rewards; those \emph{requiring holistic LLM judgment} are deferred to the preference alignment stage. This separation prevents sparse sampling noise from distorting training dynamics and avoids reward ambiguity from additional LLM-based evaluators. We add two auxiliary rewards beyond the RACE Rubrics signal:

\textbf{Citation Reward $R_{\text{cite}}$.} Let $n_{\text{gen}}$ denote the number of citations in the generated report, $n_{\text{ref}}$ the number in the reference report, and $n_{\text{valid}}$ the number of valid citations---a citation is considered valid if the textual relevance between its description and the cited passage exceeds threshold $\tau$. We define:
\begin{equation}
R_{\text{cite}} =
\begin{cases}
+0.1 & n_{\text{gen}} \geq 0.7\,n_{\text{ref}} \;\text{and}\; n_{\text{valid}} \geq 0.7\,n_{\text{ref}} \\
-0.1 & n_{\text{gen}} \geq 0.7\,n_{\text{ref}} \;\text{and}\; n_{\text{valid}} < 0.7\,n_{\text{ref}} \\
-1   & n_{\text{gen}} < 0.7\,n_{\text{ref}}
\end{cases}
\end{equation}
Sufficient citations with adequate validity receive a positive reward; sufficient citations with poor validity incur a mild penalty; insufficient citations incur a heavy penalty.

\textbf{Format Reward $R_{\text{format}}$.} We define binary violation indicators for three structural issues in the generated report:
\begin{itemize}
    \item $v_{\text{tag}} = \mathbb{1}[\text{final answer not properly enclosed in \texttt{<final\_answer>} tags}]$
    \item $v_{\text{md}} = \mathbb{1}[\text{Markdown formatting errors (list numbering, table rendering, etc.)}]$
    \item $v_{\text{ref}} = \mathbb{1}[\text{citation formatting errors}]$
\end{itemize}
Each violation is penalized independently:
\begin{equation}
R_{\text{format}} = -\bigl(v_{\text{tag}} + v_{\text{md}} + v_{\text{ref}}\bigr) \;\in\; [-3,\; 0]
\end{equation}

The overall reward is:
\begin{equation}
R_{\text{Report}} = R_{\text{RACE}} + \lambda_c R_{\text{cite}} + \lambda_f R_{\text{format}}
\end{equation}
where $R_{\text{RACE}}$ is the weighted RACE score, and $\lambda_c$, $\lambda_f$ are tunable balancing coefficients. Different base models exhibit different rates of citation deficiency and format violations during training; calibrating these coefficients to observed error rates enables fine-grained control over auxiliary reward influence on the primary optimization direction, preventing convergence disruption from over- or under-penalizing individual issues.

\paragraph{Optimization Objective.}
We adopt GRPO as the baseline optimization algorithm. While GRPO achieves stable results on dense models, its limitations are amplified in long-form generation: sequence-level policy gradients assign equal weight to responses of varying length, diluting gradient contributions from longer sequences; symmetric clipping restricts upward exploration of importance ratios, risking entropy collapse; and the KL divergence constraint further suppresses exploration. For Report-RL on dense models, we therefore adopt DAPO~\cite{yu2025dapo}.

Relative to GRPO, DAPO introduces four key improvements: (1) \textbf{token-level policy gradients}, normalizing loss by total token count to eliminate gradient imbalance across response lengths; (2) \textbf{asymmetric clipping} (clip-higher), using $\varepsilon_{\text{low}} < \varepsilon_{\text{high}}$ to provide greater upward exploration room and suppress entropy collapse; (3) \textbf{removal of the KL constraint}, enhancing free exploration in reward space; and (4) \textbf{dynamic sampling filter}, discarding groups where all samples are correct or all incorrect to ensure non-trivial advantage estimates. Letting $r_{i,t}(\theta) = {\pi_\theta(y_{i,t} \mid x, y_{i,<t})}/{\pi_{\theta_{\text{old}}}(y_{i,t} \mid x, y_{i,<t})}$, the DAPO objective is:
\begin{equation}
J_{\text{DAPO}}(\theta) = \mathbb{E}_{x \sim \mathcal{D}',\, \{y_i\} \sim \pi_{\theta_{\text{old}}}} \left[ \frac{1}{\sum_i |y_i|} \sum_i \sum_t \min\!\left( r_{i,t}(\theta)\,\hat{A}_i,\;\operatorname{clip}\!\left(r_{i,t}(\theta), 1{-}\varepsilon_{\text{low}}, 1{+}\varepsilon_{\text{high}}\right)\hat{A}_i \right) \right]
\end{equation}
where $\mathcal{D}'$ is the dynamically filtered training set and $\hat{A}_i$ are group-relative advantages.

Given the high inference cost of Qwen3-32B, we also explore Qwen3-30B-A3B (MoE) as a unified backbone. Token-level importance ratios are highly sensitive to expert routing changes under MoE sparse activation, causing the same clipping instability for DAPO as for GRPO. We therefore apply GSPO~\cite{zheng2025gspo} consistently with the Search-RL stage (Section~\ref{sec:search_rl}): its sequence-level importance ratio $s_i(\theta)$ averages over local routing changes, stabilizing the clip constraint and preventing report writing capability from degrading.

\subsection{Preference Alignment}
\label{sec:reference_align}
After SFT and RL training, the model has acquired strong deep search and report writing capabilities. However, RL reward design is primarily driven by quantifiable objective metrics and is inherently limited in addressing experience-quality issues that \emph{require holistic LLM judgment to identify and cannot be defined by simple rules}---such as tense inconsistency, unnatural tone, abrupt paragraph transitions, and conclusions inconsistent with retrieved content. The preference alignment stage aims to close this gap by systematically shifting the model's output distribution from low-quality regions toward high-quality regions aligned with human expectations.

\paragraph{Data Construction.}
To maintain consistency between training data and the model's own generation distribution, we adopt a \textbf{self-sampling} strategy: for each input $x$, we sample multiple outputs $\{y_1, \dots, y_K\}$ from the current policy and score each through a quality pipeline combining two signal types: (1) \textbf{LLM-as-Judge}, where multiple independent evaluators assess fine-grained dimensions including comprehensiveness, logical consistency, language quality, and tense accuracy; and (2) \textbf{rule-based detection}, applying hard penalties for detectable structural issues (missing citations, format violations, etc.). The two signals are weighted and combined into a quality score $s(x, y)$, which is used to partition outputs into a high-quality set $\mathcal{D}^+$ and a low-quality set $\mathcal{D}^-$.

\begin{table*}[t]
\centering
\caption{Performance on five DS benchmarks. Best results in our evaluation environment are shown in bold, and second-best results are underlined.}
\label{tab:main}
\resizebox{0.8\textwidth}{!}{%
\begin{tabular}{lccccc}
\toprule
\textbf{Model} & \makecell{\textbf{Browse}\\\textbf{Comp-ZH}} & \makecell{\textbf{Browse}\\\textbf{Comp}} & \makecell{\textbf{xbench}\\\textbf{-DS}} & \makecell{\textbf{GAIA}\\\textbf{-DS}}  & \makecell{\textbf{Wide}\\\textbf{Search}} \\
\midrule
\multicolumn{6}{c}{\textit{Large-Scale Foundation Models}} \\
\midrule
GLM-4.6 \cite{zai2025glm46}               & \underline{45.1} & \textbf{49.5} & \underline{73.0}   & 52.6    & 43.1 \\
Kimi K2 \cite{team2025kimi}               & 28.8 & 14.1 & 50.0 & 57.7 & 54.4 \\
DeepSeek R1 \cite{deepseek2025r1}               & 34.6 & 14.1 & 50.0 & 57.7  & 44.3 \\
Qwen3-235B \cite{yang2025qwen3}      & 31.1 & 21.7 & 57.0 & 63.1  & \underline{46.4} \\
\midrule
\multicolumn{6}{c}{\textit{Comparable-Scale Agent Models}} \\
\midrule
WebDancer-32B \cite{wu2025webdancer}  & 25.3 & 10.5  & 11.0 & 63.1  & 39.7 \\
WebSailor-32B \cite{websailor2025}            & 25.6 & 14.8  & 46.0 & 50.5  & 40.3 \\
WebShaper-32B \cite{tao2025webshaperagenticallydatasynthesizing}             & 28.0 & 33.5  & 53.0 & 54.4  & 35.2 \\          
MiroThinker-v1.5-30B-A3B \cite{team2025mirothinker} & 31.9  & 30.4  & 5.0  & 23.3  & 37.9 \\
OpenSeeker-30B-A3B\cite{du2026openseekerdemocratizingfrontiersearch}        & 26.4 & 12.9 & 48.5 & 46.7  & 36.4 \\
Tongyi-DR-30B-A3B \cite{team2025tongyi}        & 43.2 & 40.7 & 69.0 & \underline{68.9}  & 41.7 \\
\midrule
\multicolumn{6}{c}{\textit{Our Agent Models}} \\
\midrule
MindDR-v1.0-32B                          & 28.4 & 18.6  & 13.3 & 50.3  & 41.3 \\
\rowcolor{gray!10}
MindDR-v1.5-32B                          & 35.6 & 31.8 & 64.0 & 67.1 & \textbf{46.5}  \\
\rowcolor{gray!10}
MindDR-v1.5-30B-A3B                      & \textbf{45.7} & \underline{42.8} & \textbf{75.0} & \textbf{70.9} & 44.0 \\
\bottomrule
\end{tabular}%
}
\end{table*}

\paragraph{Training Methods.}
The alignment stage employs two complementary methods---DPO~\cite{rafailov2023dpo} and Self-SFT~\cite{chen2025selfsft}---driving distribution shift from the perspectives of preference contrast and behavior cloning, respectively.

\textbf{DPO.} We construct a preference dataset $\mathcal{D}_{\text{pref}} = \{(x, y^+, y^-)\}$ from high- and low-scoring output pairs for the same input, where $y^+ \in \mathcal{D}^+$ and $y^- \in \mathcal{D}^-$. DPO directly optimizes the log-probability margin of the policy relative to a reference policy, without requiring an explicit reward model:
\begin{equation}
\mathcal{L}_{\text{DPO}}(\theta) = -\mathbb{E}_{(x,\,y^+,\,y^-) \sim \mathcal{D}_{\text{pref}}} \left[ \log \sigma \!\left( \beta \log \frac{\pi_\theta(y^+ \mid x)}{\pi_{\text{ref}}(y^+ \mid x)} - \beta \log \frac{\pi_\theta(y^- \mid x)}{\pi_{\text{ref}}(y^- \mid x)} \right) \right]
\end{equation}
where $\pi_{\text{ref}}$ is the reference policy at the start of the alignment stage and $\beta$ controls preference strength.

\textbf{Self-SFT.} We directly fine-tune on samples in $\mathcal{D}^+$ to reinforce the model's ability to reproduce high-quality output patterns:
\begin{equation}
\mathcal{L}_{\text{Self-SFT}}(\theta) = -\mathbb{E}_{(x,\,y^+) \sim \mathcal{D}^+} \left[ \log \pi_\theta(y^+ \mid x) \right]
\end{equation}
Self-SFT can be understood as \textbf{reinforcement on static data}: analogous to RL's mechanism of driving policy updates through online sampling, Self-SFT uses high-quality self-sampled data as supervision signals to improve output quality through static iteration while preserving on-policy characteristics. Compared to online RL, it incurs lower computational cost and exhibits more stable convergence, making it a suitable complement to the alignment stage.

Both methods build training data exclusively from the model's own samples, constraining distribution adjustment to the model's existing output space. This prevents the policy from being pulled toward unexplored regions, preserving the search, reasoning, and writing capabilities accumulated during SFT and RL, while continuously shifting outputs toward quality levels aligned with human expectations.

\section{Main Results}
\label{sec:results}
In this section, we evaluate the performance of MindDR across various deep search and deep research tasks. We first introduce the experimental setup, detailing the compared baselines and the evaluation benchmarks. We then present the main results in three parts: DeepSearch benchmark performance, DeepResearch benchmark performance, and the final gains from preference alignment.

\subsection{Experimental Setup}

\paragraph{Evaluated Models.} We evaluate our models against a diverse set of strong baselines, including both large-scale foundation models and comparable-scale agent systems:
\begin{itemize}[leftmargin=1.5em,itemsep=2pt]
    \item \textbf{Large-Scale Foundation Models:} We compare against leading proprietary and open-source models, including Gemini 3.1~\cite{googledeepmind2026gemini31pro}, Gemini 2.5 Pro~\cite{comanici2025gemini25}, GLM-4.6~\cite{zai2025glm46}, Kimi K2~\cite{team2025kimi}, DeepSeek R1~\cite{deepseek2025r1}, Doubao, and Qwen3-235B~\cite{yang2025qwen3}.
    \item \textbf{Comparable-Scale Agent Systems:} We evaluate against leading open-source agent models around the 30B parameter scale, including WebDancer-32B~\cite{wu2025webdancer}, WebSailor-32B~\cite{websailor2025}, WebShaper-32B~\cite{tao2025webshaperagenticallydatasynthesizing}, MiroThinker-v1.5-30B-A3B~\cite{team2025mirothinker}, OpenSeeker-30B-A3B~\cite{du2026openseekerdemocratizingfrontiersearch}, Tongyi-DR-30B-A3B~\cite{team2025tongyi}.
    \item \textbf{MindDR Variants:} We also evaluate two versions of our system: MindDR-v1.0, which is trained using only the Reinforcement Fine-Tuning (RFT) stage, and MindDR-v1.5, the fully upgraded version that incorporates the entire RL pipeline (Search-RL, Report-RL, and Preference Alignment). In addition, we have MindDR-v1.5-32B to represent MindDR-v1.5 with Qwen3-32B as the base model and MindDR-v1.5-30B-A3B with Qwen3-30B-A3B as the base model. 
\end{itemize}

\paragraph{Benchmarks.} We comprehensively evaluate the systems across two distinct categories of benchmarks:
\begin{itemize}[leftmargin=1.5em,itemsep=2pt]
    \item \textbf{DeepSearch (DS) Benchmarks:} We measure multi-step information retrieval and reasoning capabilities using BrowseComp-ZH~\cite{zhou2025browsecompzh}, BrowseComp~\cite{wei2025browsecomp}, xbench-DS~\cite{chen2025xbench}, GAIA-DS~\cite{grégoire2023gaia}, and WideSearch~\cite{wong2025widesearch}. These benchmarks require agents to autonomously navigate web environments and extract accurate answers for complex queries.
    \item \textbf{DeepResearch (DR) Benchmarks:} We assess the ability to generate comprehensive, structured, and human-aligned long-form reports on DeepResearch Bench~\cite{du2025deepresearchbench} and our proposed \textbf{MindDR Bench}. Evaluation is conducted using RACE rubrics covering comprehensiveness, insight, instruction following, and readability alongside user experience metrics such as citation accuracy, table format and temporal error.
\end{itemize}

\begin{figure}[t]
    \centering
    \includegraphics[width=0.9\columnwidth]{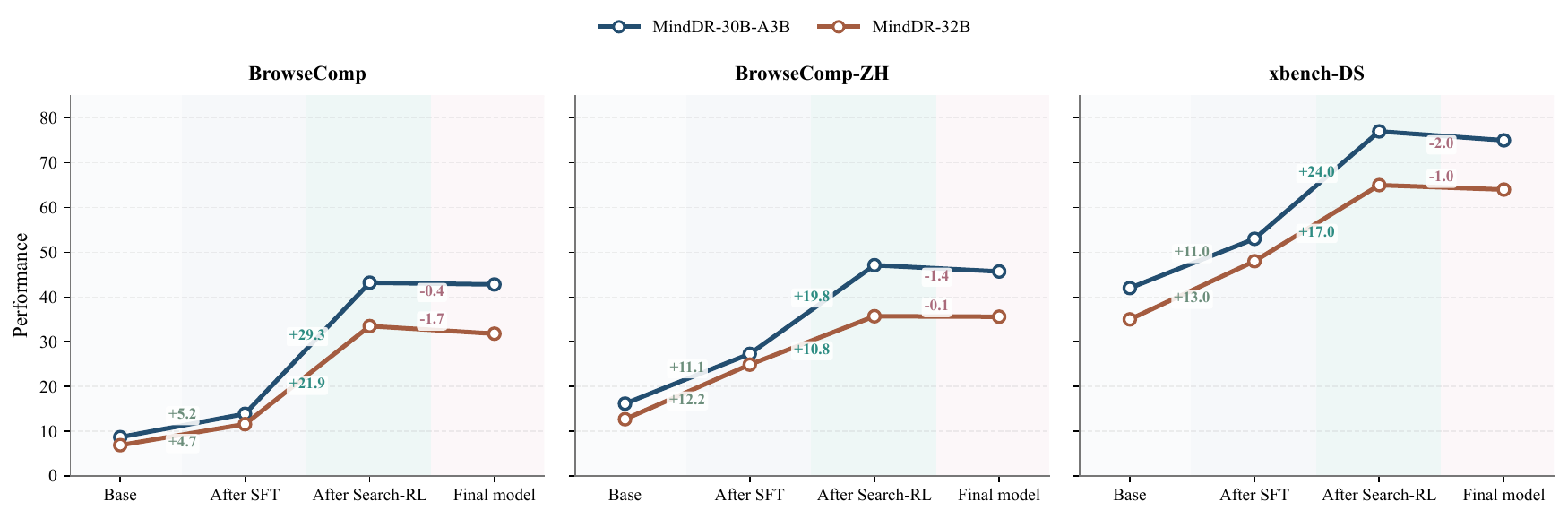}
    \caption{Stage-wise DS benchmark performance from the base model to \textsc{SFT}, \textsc{Search-RL}, and the final model. \textsc{Search-RL} consistently delivers the largest gains across all three benchmarks and both model sizes, while the final stage introduces only minor regressions, indicating a small trade-off in search performance.}
    \label{fig:searchrl_gains}
\end{figure}

\begin{table}[h]
\centering
\caption{Performance on MindDR Bench on RACE score and Citation Accuracy. Best results are shown in bold, and second-best results are underlined. \textbf{Comp.}: Comprehensiveness; \textbf{Inst.}: Instruction Following; \textbf{Read.}: Readability; \textbf{C.Acc.}: Citation Accuracy.}
\label{tab:mindr500_results}
\resizebox{0.85\columnwidth}{!}{%
\begin{tabular}{lcccccc}
\toprule
\textbf{Model} & \textbf{RACE} & \textbf{Comp.} & \textbf{Insight} & \textbf{Inst.} & \textbf{Read.} & \textbf{C.Acc.} \\
\midrule
Gemini 3.1       & \underline{49.65} & \underline{49.83} & \underline{50.75} & 49.32          & 46.89          & 77.20\%              \\
Gemini 2.5 Pro   & 48.34          & 47.56          & 46.88           & \underline{49.47} & \underline{50.23}          & \underline{81.86\%}  \\
Doubao           & 46.25          & 48.21          & 40.88           & 48.71         & 48.18          & 68.43\%              \\
Kimi             & 45.20          & 46.08          & 40.60           & 47.94         & 47.53          & 75.01\%              \\
Qwen             & 45.07          & 45.55          & 39.94           & 48.34         & 48.01          & 76.42\%              \\
\midrule
MindDR-v1.0      & 44.33          & 44.72          & 39.49           & 47.58         & 47.66          & \textbf{82.14\%}     \\
\rowcolor{gray!10}
MindDR-v1.5      & \textbf{51.77} & \textbf{52.17} & \textbf{51.77} & \textbf{50.55} & \textbf{52.18} & 80.25\%                    \\
\bottomrule
\end{tabular}%
}
\end{table}

\subsection{Overall Performance}
\label{sec:overall_results}

Table~\ref{tab:main} summarizes the overall DeepSearch performance. MindDR-v1.5-30B-A3B establishes a strong DS frontier among open-source agent-style systems in our evaluation environment, achieving the best results on BrowseComp-ZH, BrowseComp, xbench-DS, and GAIA-DS. MindDR-v1.5-32B achieves the best WideSearch result, indicating that the gains generalize across backbones rather than being specific to a single checkpoint. Overall, the final MindDR models close or surpass the gap to stronger foundation-style baselines while clearly outperforming comparable-scale open agent systems.

\begin{figure}[t]
    \centering
    \includegraphics[width=0.96\linewidth]{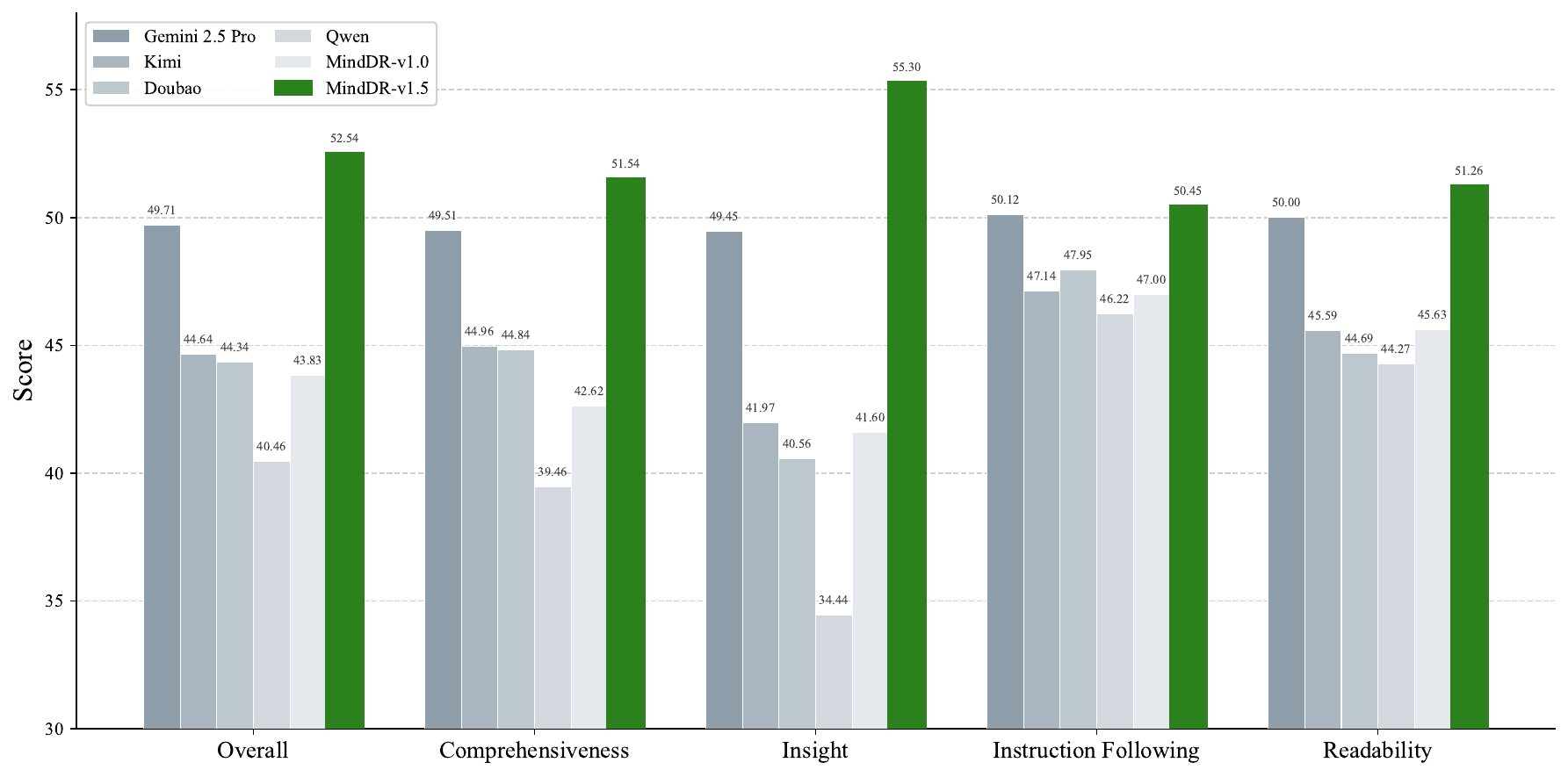}
    \caption{Comparison with mainstream DR systems on the public DeepResearch-Benchmark leaderboard. Each group of bars represents a evaluation dimension, and the scores are annotated above the corresponding bars. MindDR-v1.5 (green) achieves the highest scores across all five metrics.}
    \label{fig:dense_report_rl}
\end{figure}

We next evaluate DR quality using RACE and its subdimensions on MindDR Bench. Fig.~\ref{fig:dense_report_rl} and Table~\ref{tab:mindr500_results} show that MindDR-v1.5 moves into the top tier of DR systems. The gains are broad rather than metric-specific: MindDR-v1.5 leads on overall RACE and on the main report-quality dimensions, including comprehensiveness, insight, instruction following, and readability. At the product level, the main remaining exception is citation accuracy, where MindDR-v1.0 remains slightly stronger, suggesting that report-quality optimization and citation-faithfulness optimization are related but not identical objectives.


The DS and DR results together indicate that MindDR achieves a favorable balance between strong search performance and high-quality report generation. The system is therefore competitive not only as a search agent, but also as a complete deep research system.

\begin{table}[!h]
\centering
\caption{Report-RL ablation results evaluated on MindDR Bench. Higher RACE, tag format, citation format, and BrowseComp-ZH are better; lower table error is better.}
\label{tab:reportrl_ablation}
\resizebox{0.95\columnwidth}{!}{%
\begin{tabular}{lccccc}
\toprule
\textbf{Model} & \textbf{RACE} & \textbf{Tag Format} & \textbf{Citation Format} & \textbf{Table Error} & \textbf{BrowseComp-ZH} \\
\midrule
Qwen3-30B-A3B + Search-RL & 39.54          & 69\%          & 78\%          & 0.41\%        & \textbf{47.1\%} \\
\quad + Report-RL (GRPO)  & 41.68          & 91\%            & 98\%            & 0.85\%            & 29.1\%          \\
\quad + Report-RL (DAPO)  & 43.27          & 95\%            & 99\%            & 0.65\%           & 43.3\%          \\
\quad + Report-RL (GSPO)  & 44.05 & \textbf{99\%} & \textbf{99\%} & 0.50\%        & 45.7\%          \\
\midrule
Qwen3-32B +SFT + Search-RL & 43.32          & 96\%   & 97\%     & 0.41\%        & 35.7\% \\
\quad + Report-RL (GRPO)&     48.06   & 97\%   & 97\%     &  2.60\%        &  28.6\% \\
\quad + Report-RL (DAPO)  & \textbf{48.82}   & \textbf{99\%} & \textbf{99\%}    & 2.70\%      & 35.6\%          \\
\bottomrule
\end{tabular}%
}
\end{table}

\subsection{Detailed Analysis}
\label{sec:data_efficiency}

We now turn to a more detailed analysis of the training stages, with particular attention to the trade-off between DS capability, DR quality, and final product-level refinement.

Table~\ref{tab:reportrl_ablation} shows that the main challenge in DR optimization is not whether Report-RL improves report quality---it does---but how much prior DS capability is preserved while doing so. Sequence-level methods such as GSPO and DAPO provide the best balance in this regard: they improve report quality and formatting substantially while keeping search regression small, whereas GRPO causes noticeably larger DS degradation.

\begin{table}[t]
\centering
\caption{Ablation results of long-form and mixed-form inputs for Report-RL evaluated on MindDR Bench.}
\label{tab:long-short-form}
\resizebox{0.8\columnwidth}{!}{%
\begin{tabular}{lccccc}
\toprule
\textbf{Model} & \textbf{RACE} & \textbf{Comp.} & \textbf{Insight} & \textbf{Inst.} & \textbf{Read.} \\
\midrule
Qwen3-32B + SFT + Search-RL                        & 43.32 & 44.30 & 37.93 & \textbf{46.96} & 45.89 \\
\quad + Report-RL (DAPO), Long Only      & 48.82 & 49.10 & 47.68 & 49.42 & 49.56 \\
\quad + Report-RL (DAPO), Long + Short   & \textbf{50.60} & \textbf{51.08} & \textbf{50.96} & 49.67 & \textbf{49.66} \\
\bottomrule
\end{tabular}%
}
\end{table}

\paragraph{Effects of Long-form and Short-form Queries.}
Starting from an SFT checkpoint, we apply DAPO~\cite{yu2025dapo} with RACE Rubrics as the reward. Table~\ref{tab:long-short-form} compares Report-RL trained on long-form-only data against a mixture of long-form and short-form data. Adding short-form data yields a consistent gain across all metrics: RACE improves from 48.82 to 50.60, with notable increases in Comprehensiveness (49.10$\to$51.08) and Insight (47.68$\to$50.96). This suggests that mixing compact supervision into long-form RL training provides a stronger optimization signal and helps the model generalize to both response lengths.

\begin{table}[h]
\centering
\caption{Effect of phase 4 preference alignment on DeepResearch Bench score and report-quality error metrics. DPO targets table and temporal correctness, while Self-SFT further improves overall writing consistency.}
\label{tab:table_reward_results}
\small
\setlength{\tabcolsep}{3pt}
\resizebox{0.92\linewidth}{!}{%
\begin{tabular}{lccccc}
\toprule
\textbf{Model} & \makecell{\textbf{DeepResearch}\\\textbf{Bench}} & \makecell{\textbf{Table}\\\textbf{Error}} & \makecell{\textbf{Post-proc.}\\\textbf{Table Error}} & \makecell{\textbf{Temporal}\\\textbf{Error}} & \makecell{\textbf{Expression / Logic}\\\textbf{Issue Rate}} \\
\midrule
MindDR-v1.5-ReportRL & 50.06 & 2.70\% & 1.35\% & 6.2\% & 1.8\% \\
\quad + DPO & 50.07 & 1.22\% & 0.16\% & 2.0\% & 1.8\% \\
\quad + Self-SFT & \textbf{51.77} & \textbf{1.22\%} & \textbf{0.16\%} & \textbf{2.0\%} & \textbf{0.3\%} \\
\bottomrule
\end{tabular}%
}
\end{table}

\paragraph{Quality Refinement by preference alignment.}
Even after Report-RL, rubric-based evaluation reveals residual defects that are only weakly captured by scalar DR rewards, including table-information misalignment, temporal-expression errors, paragraph-level logical discontinuities, and language inconsistency. We therefore apply a final refinement stage composed of DPO and Self-SFT. DPO targets the most structured and objectively judgeable issues---table correctness and temporal correctness---using a curated 1.8K temporal dataset and a 2.8K table-repair dataset, while Self-SFT on 4.3K high-quality self-sampled reports improves coherence and stylistic consistency.

\begin{table}[h]
\centering
\caption{Examples of temporally correct and temporally incorrect statements in generated reports.}
\label{tab:temporal_examples}
\resizebox{0.95\columnwidth}{!}{%
\begin{tabular}{p{2cm} p{6cm} p{5cm}}
\toprule
\textbf{Type} & \textbf{Example} & \textbf{Explanation} \\
\midrule
Correct &
1. ``According to XX organization in 2024, the global VR/AR training market is expected to reach tens of billions by 2025.'' \newline
2. ``The global robotics market is expected to reach hundreds of billions by 2027.'' &
1. Prediction about a past time point with explicit attribution. \newline
2. Prediction about a future time point, consistent with temporal logic. \\
\midrule
Incorrect &
``By 2025, the global VR/AR training market will reach tens of billions.'' &
Prediction about a past time point without attribution. \\
\bottomrule
\end{tabular}
}
\end{table}

Table~\ref{tab:table_reward_results} shows a clear division of labor between the two refinement steps. DPO sharply reduces structured factual and formatting errors, especially table and temporal errors, while leaving the aggregate DR benchmark essentially unchanged. Self-SFT then improves higher-level expression and coherence, reducing the expression/logic issue rate further and yielding an additional gain on the benchmark itself. In combination, the two stages improve report correctness and presentation without sacrificing the performance established by Report-RL.

We define a temporal error as a predictive statement about a past time point that lacks explicit attribution to a source. 

\begin{table}[!h]
\centering
\begin{minipage}[h]{0.43\linewidth}
\caption{Temporal error rates on MindDR Bench. Lower is better.}
\label{tab:temporal_error}
\small
\setlength{\tabcolsep}{3pt}
\resizebox{0.95\linewidth}{!}{%
\begin{tabular}{lc}
\toprule
\textbf{Model} & \textbf{Temporal Error Rate}\\
\midrule
MindDR-v1.5 & \textbf{2.0\%} \\
Kimi        & 3.2\% \\
Qwen3       & 8.2\% \\
Gemini      & 10.2\% \\
MindDR-v1.0 & 11.4\% \\
Doubao      & 14.0\% \\
\bottomrule
\end{tabular}%
}
\end{minipage}\hfill
\begin{minipage}[h]{0.52\linewidth}
\vspace{0pt}
\small
Table~\ref{tab:temporal_examples} illustrates this distinction, and the temporal error comparison in Table~\ref{tab:temporal_error} shows that MindDR-v1.5 achieves the lowest temporal error rate among all compared systems. This indicates that the final refinement stage corrects a report defect that remains common even in strong industrial systems and helps turn a strong model into a product-quality system.

MindDR-v1.5 achieves the lowest temporal error rate among all compared systems at 2.0\%, substantially improving over MindDR-v1.0 and outperforming all external baselines. This highlights the effectiveness of the final refinement stage in correcting a subtle but important report-quality defect.
\end{minipage}
\end{table}

\begin{figure}[h]
    \centering
    \includegraphics[width=0.95\linewidth]{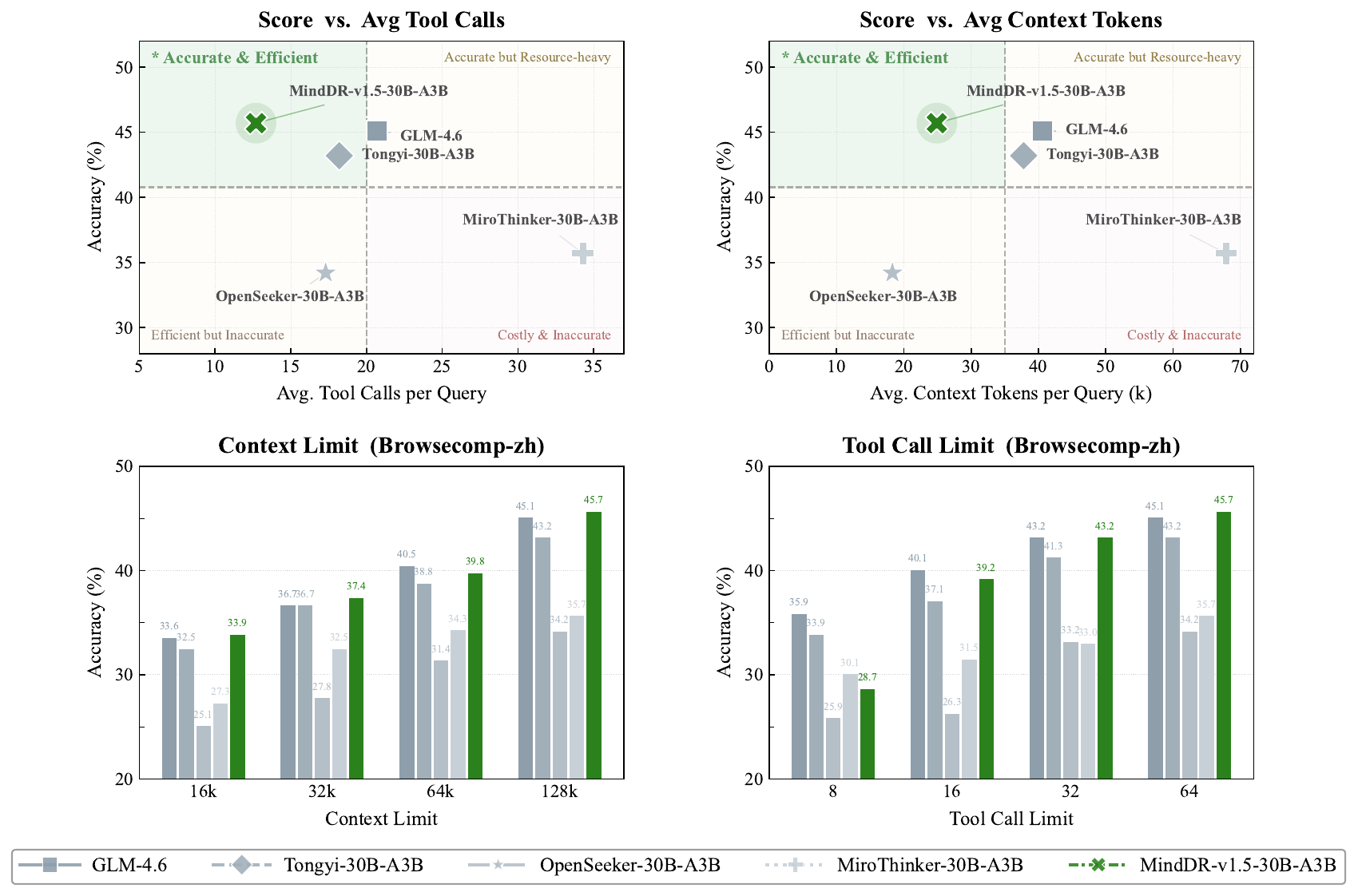}
    \caption{Efficiency and scalability analysis on BrowseComp-ZH. \textbf{Top row:} quadrant plots of accuracy versus average tool calls (left) and context tokens (right) per query. \textbf{Bottom row:} accuracy under varying context-length limits (left) and tool-call limits (right). MindDR-v1.5-30B-A3B consistently occupies the \emph{Accurate \& Efficient} region and maintains leading performance across all budget settings among comparable 30B systems.}
    \label{fig:bench_effi}
\end{figure}

\paragraph{Training and Test-time Efficiency.}
Fig.~\ref{fig:bench_effi} examines inference efficiency on BrowseComp-ZH from four complementary perspectives: accuracy versus average tool calls (upper-left) and average context tokens (upper-right), and performance under varying context limits (lower-left) and tool-call limits (lower-right). The upper scatter plots adopt a quadrant view where the top-left region denotes the ideal \emph{Accurate \& Efficient} zone.

The five compared systems exhibit distinct efficiency--accuracy trade-offs. OpenSeeker-30B-A3B is the most resource-frugal yet its accuracy lags noticeably. Miro-30B-A3B sits at the opposite extreme, incurring the highest context and tool-call consumption while delivering limited accuracy gains. GLM-4.6, a leading proprietary model, achieves competitive scores at the cost of substantially greater resource usage. In contrast, MindDR-v1.5-30B-A3B falls squarely within the \emph{Accurate \& Efficient} quadrant, attaining the highest BrowseComp-ZH score (45.7) while requiring the fewest average context tokens and tool calls among top-performing systems.

The lower bar charts confirm that this advantage is robust across operational constraints. Under context limits from 16k to 128k and tool-call limits from 8 to 64, MindDR-v1.5-30B-A3B consistently matches or surpasses all compared models. Although it starts modestly under the most restrictive tool-call budget, it scales rapidly and achieves the best accuracy at standard and permissive settings, indicating that its deep-search strength reflects genuinely higher per-step retrieval efficiency rather than simply larger contexts or more aggressive tool use.

More broadly, these results support the efficiency claim of the overall pipeline. A key difference from many industry deep-research systems is that MindDR removes the midtraining stage that is often used to inject tool use, strategy selection, and reflection behaviors into the base model, sometimes at a scale exceeding 150B tokens. This design also yields a clear training-efficiency advantage relative to our previous-generation system. A single MindDR 1.0 training run used 280K high-quality RFT samples, corresponding to approximately 3.6B tokens, together with about 15K GPU card-hours. By contrast, MindDR 1.5 uses only about 0.18B SFT tokens and 0.85B RL sampling tokens, for a total of roughly 1.03B training-related tokens, and requires only about 6K GPU card-hours. Instead of relying on a large intermediate training phase followed by end-to-end RL, MindDR 1.5 adopts a staged optimization design: the system first establishes cold-start behavior with SFT, then separately optimizes search capability with Search-RL, report generation with Report-RL, and final product quality with preference alignment. This decomposition substantially reduces training resource requirements while yielding stronger downstream performance. The results therefore suggest that MindDR's gains arise less from brute-force scaling and more from replacing monolithic midtraining-plus-end-to-end optimization with a more efficient staged training and reward design.

\section{Discussion and Conclusion}
\label{sec:conclusion}
\subsection{Limitations}

\paragraph{Context Management.}
While our progressive length generalization strategy achieves 94\% format correctness at 128K context, scaling to even longer contexts for extremely complex research tasks remains an open challenge. Developing more effective context management strategies---such as hierarchical memory, selective context compression, or adaptive attention mechanisms---that allow agents to maintain focus on the most relevant evidence while operating over very long horizons is an important direction for future work.

\paragraph{Evaluation coverage.}
Current evaluation primarily relies on the RACE framework and factual accuracy metrics. However, deeper aspects of research quality---such as methodological soundness, argument novelty, and appropriate hedging of uncertain claims---are difficult to capture with automated rubrics and would benefit from more nuanced human evaluation protocols.

\subsection{Conclusion}

We presented MindDR, a cost-effective, open multi-agent framework designed to address the fundamental bottleneck of deep research agents: achieving top-tier performance and excellent user experience without relying on prohibitive training and inference costs. By avoiding computationally expensive continual pre-training and adopting a highly targeted optimization strategy, MindDR demonstrates that \textasciitilde30B-parameter models can match or surpass the deep research capabilities of much larger foundation models.

MindDR tackles the cost-performance trade-off through both inference-stage decomposition and training-stage targeted optimization. At inference, the three-agent architecture (Planning, DeepSearch, and Report Agents) coordinates via an Extended Chain-of-Thought mechanism to parallelize search and isolate contexts, naturally alleviating long-context burdens. At training, the staged pipeline progresses from SFT to Search-RL (explicitly optimizing search efficiency to reduce redundant token consumption), then to Report-RL (resolving information conflicts for long-form generation), and finally to preference alignment (correcting residual formatting and temporal defects). 

Empirically, MindDR achieves strong results on both DeepSearch (DS) and DeepResearch (DR) evaluations. On DS benchmarks, MindDR-v1.5-30B-A3B attains the best results among open-source agent-style systems on BrowseComp-ZH, BrowseComp, xbench-DS, and GAIA-DS, while maintaining its advantage under varying context-length and tool-call constraints. Furthermore, we introduced MindDR Bench alongside a comprehensive multi-dimensional evaluation system, moving beyond single-metric assessments to systematically evaluate both content quality and presentation format. On this benchmark, MindDR reaches a state-of-the-art RACE score of 51.8, leading across comprehensiveness, insight, instruction following, and readability.

\section{Appendix}
\label{sec:appendix}
\subsection{RACE Rubrics Example}
\label{sec:race_rubrics_example}

Below is an illustrative example of the RACE Rubrics generated for a specific query. Each rubric contains four evaluation dimensions---comprehensiveness, insight, instruction following, and readability---with multiple fine-grained criteria, explanations, and importance weights. The scoring model uses these rubrics to produce per-dimension scores that serve as reward signals during Report-RL training. For brevity, we show a representative subset of criteria per dimension.

\begin{tcolorbox}[enhanced, breakable, colback=gray!3, colframe=gray!60, boxrule=0.4pt, arc=1.5mm, title={\small\bfseries RACE Rubrics Example (Abridged)}, fontupper=\ttfamily\tiny, left=2mm, right=2mm, top=2mm, bottom=2mm, before upper=\raggedright]
\{ \\
\quad "Query": "How has the release of the domestic LLM DeepSeek triggered educational reform?", \\
\quad "dimension\_weight": \{ \\
\quad\quad "comprehensiveness": 0.29, "instruction\_following": 0.10, \\
\quad\quad "insight": 0.44, "readability": 0.17 \\
\quad \}, \\
\quad "criterions": \{ \\
\quad\quad "comprehensiveness": [ \\
\quad\quad\quad \{ "criterion": "Completeness of DeepSeek model and release event information", \\
\quad\quad\quad\quad "explanation": "Whether the article comprehensively introduces DeepSeek's core \\
\quad\quad\quad\quad information, including R\&D institution, release timeline, key technical parameters, \\
\quad\quad\quad\quad core capabilities, and open-source/API access strategy.", \\
\quad\quad\quad\quad "weight": 0.6 \}, \\
\quad\quad\quad \{ "criterion": "Coverage of specific educational application scenarios", \\
\quad\quad\quad\quad "explanation": "Whether the article enumerates concrete applications such as \\
\quad\quad\quad\quad intelligent tutoring, AI teaching assistants, content generation, and assessment.", \\
\quad\quad\quad\quad "weight": 0.8 \}, \\
\quad\quad\quad \{ "criterion": "Identification of challenges, risks, and ethical issues", \\
\quad\quad\quad\quad "explanation": "Whether the article identifies issues including academic integrity, \\
\quad\quad\quad\quad data privacy, educational equity, and over-reliance on AI.", \\
\quad\quad\quad\quad "weight": 0.7 \} \\
\quad\quad ], \\
\quad\quad "insight": [ \\
\quad\quad\quad \{ "criterion": "Depth of reform mechanism analysis", \\
\quad\quad\quad\quad "explanation": "Whether the article deeply analyzes the core transmission mechanism \\
\quad\quad\quad\quad rather than merely listing application scenarios.", "weight": 1.0 \}, \\
\quad\quad\quad \{ "criterion": "Critical thinking and deep risk foresight", \\
\quad\quad\quad\quad "explanation": "Whether the article demonstrates balanced and critical perspectives.", \\
\quad\quad\quad\quad "weight": 0.6 \} \\
\quad\quad ], \\
\quad\quad "instruction\_following": [ \\
\quad\quad\quad \{ "criterion": "Subject focus on DeepSeek", "weight": 1.2 \}, \\
\quad\quad\quad \{ "criterion": "Scope strictly limited to education", "weight": 1.2 \} \\
\quad\quad ], \\
\quad\quad "readability": [ \\
\quad\quad\quad \{ "criterion": "Macro structure and logical framework", "weight": 0.8 \}, \\
\quad\quad\quad \{ "criterion": "Language accuracy and fluency", "weight": 0.6 \} \\
\quad\quad ] \\
\quad \} \\
\}
\end{tcolorbox}

\subsection{Short-form Data Synthesis Prompts}
\label{sec:short_text_prompt}

\begin{tcolorbox}[enhanced, breakable, colback=gray!3, colframe=gray!60, boxrule=0.4pt, arc=1.5mm, title={\small\bfseries Inference Prompt for Report Generation}, fontupper=\small\ttfamily, left=2mm, right=2mm, top=2mm, bottom=2mm, before upper=\raggedright]
\textbf{System:} You are a professional report generation agent, skilled at writing well-organized and insightful professional reports. \\[4pt]
\#\# Report Quality Requirements \\
- The report should be complete and comprehensive, without omitting key information. \\
- The report should have sectional conclusions and a final summary for easy comprehension. \\
- Sections listing facts or statistics should use tables for readability. \\[6pt]
\textbf{User:} Based on the following user QUERY, write a report. User Query: \{query\}. Generate the report directly.
\end{tcolorbox}

\vspace{4pt}

\begin{tcolorbox}[enhanced, breakable, colback=gray!3, colframe=gray!60, boxrule=0.4pt, arc=1.5mm, title={\small\bfseries Reference Report Synthesis Prompt}, fontupper=\small\ttfamily, left=2mm, right=2mm, top=2mm, bottom=2mm, before upper=\raggedright]
\textbf{System:} [Same system prompt as above] \\[6pt]
\textbf{User:} Based on the following user QUERY, considering the following dimensions \{criterions\}, write a report. User Query: \{query\}. Generate the report directly.
\end{tcolorbox}

\subsection{Scoring Model Prompt}
\label{sec:scoring_prompt}

The scoring model evaluates two articles (the generated report and a reference report) using the RACE Rubrics. The prompt template instructs the model to act as a rigorous evaluation expert, performing criterion-by-criterion comparative analysis with scores on a 0--10 scale.

\begin{tcolorbox}[enhanced, breakable, colback=gray!3, colframe=gray!60, boxrule=0.4pt, arc=1.5mm, title={\small\bfseries Scoring Model Prompt Template}, fontupper=\small\ttfamily, left=2mm, right=2mm, top=2mm, bottom=2mm, before upper=\raggedright]
\textbf{System:} You are a rigorous, meticulous, and objective research article evaluation expert. You excel at comparing two articles written for the same task based on specific evaluation criteria, providing precise scores and clear justifications. \\[6pt]
\textbf{User:} \\[2pt]
\textbf{**Task Background**} \\
A deep research task requires evaluating two articles across four dimensions: comprehensiveness, insight, instruction following, and readability. \\[2pt]
<task>\{task\_prompt\}</task> \\[4pt]
\textbf{**Articles to Evaluate**} \\
<article\_1>\{article\_1\}</article\_1> \\
<article\_2>\{article\_2\}</article\_2> \\[4pt]
\textbf{**Evaluation Criteria**} \\
<criteria\_list>\{criteria\_list\}</criteria\_list> \\[4pt]
\textbf{**Scoring Rules**} \\
\quad 0--2: Very poor. Almost completely fails to meet the criterion. \\
\quad 2--4: Poor. Minimally meets the criterion with obvious deficiencies. \\
\quad 4--6: Average. Basically meets the criterion. \\
\quad 6--8: Good. Largely meets the criterion with notable strengths. \\
\quad 8--10: Excellent. Fully or exceeds expectations. \\[4pt]
\textbf{**Output Format**} \\
\{ "comprehensiveness": [ \{ "criterion": "[text]", "analysis": "[analysis]", \\
\quad "article\_1\_score": [0-10], "article\_2\_score": [0-10] \}, ... ], \\
\quad "insight": [...], "instruction\_following": [...], "readability": [...] \}
\end{tcolorbox}

\subsection{Temporal Tense Error Detection}
\label{sec:tense_detection}

The temporal tense error detection pipeline consists of two stages:

\paragraph{Stage 1: Regex-based extraction.}
We use a rule-based extractor to identify sentences containing predictive temporal expressions. The extractor matches date patterns (e.g., ``by 2025'', ``in 2024Q3'') co-occurring with future-tense keywords (e.g., ``is expected to'', ``will'', ``is projected to'') within a bounded text window. Sentences where the predicted date falls before the current date are flagged as candidates for tense errors.

\paragraph{Stage 2: LLM-based institution verification.}
Each flagged sentence, together with its surrounding context (1--2 sentences before and after), is fed to an LLM with the following verification prompt:

\begin{tcolorbox}[enhanced, breakable, colback=gray!3, colframe=gray!60, boxrule=0.4pt, arc=1.5mm, title={\small\bfseries Temporal Verification Prompt}, fontupper=\small\ttfamily, left=2mm, right=2mm, top=2mm, bottom=2mm, before upper=\raggedright]
\textbf{System:} You are a rigorous fact-checking assistant. Based on the context, objectively determine whether an explicit forecasting entity is mentioned. Output only the specified JSON format. \\[6pt]
\textbf{User:} Context: \{context\} \\[2pt]
Please determine: within 1--2 sentences of "\{matched\_text\}", is there an explicit mention of the \textbf{forecasting institution} that made this prediction? \\[4pt]
"Forecasting institution" includes: \\
- Named institutions (e.g., IMF, Goldman Sachs) \\
- Generic but explicit third parties (e.g., "a research institute") \\
- The company itself describing its own plans (e.g., "the company expects") \\[4pt]
Output: \{"result": "yes/no", "detail": "brief explanation"\}
\end{tcolorbox}

\vspace{4pt}
A sentence is classified as a tense error if and only if (i) it contains a prediction about a past time point and (ii) no forecasting institution is identified in the surrounding context.

\newpage
\section*{Contributions}
\addcontentsline{toc}{section}{Contributions}

\textbf{Project Lead} \\
Sheng Yang

\vspace{4pt}
\textbf{Core Contributors} \\
Biao Wang, Haozhi Xie, Heyang Xu, Liping Wang, Shirui Zhang, Shuai Wang, Sirui Miao, Tiankuo Xu, Xuefeng Hao, Ying Liu, Yingjie Feng, Yuchen Liu, Yuhang Wu, Zhengxin Yu, Zhuo Liu

\vspace{4pt}
\textbf{Contributors} \\
Bin Huang, Dong Wu, Handong Cui, He Cao, Jiabang He, Jiajun Yang, Jialu Chen, Jiqing Zhan, Li Gong, Lian Wen, Qingfeng Cai, Xiaobo Liu, Yuan Xue, Yun Zhu


\vspace{4pt}
\textbf{Sponsors} \\
Xiaofei Gou, Wei Chen
\newpage

\bibliographystyle{plainnat}
\bibliography{references}

\end{document}